\documentclass[lettersize,journal]{IEEEtran}
\usepackage{amsmath,amsfonts}
\usepackage{algorithmic}
\usepackage{algorithm}
\usepackage{array}
\usepackage[caption=false,font=normalsize,labelfont=sf,textfont=sf]{subfig}
\usepackage{textcomp}
\usepackage{stfloats}
\usepackage{url}
\usepackage{verbatim}
\usepackage{graphicx}
\usepackage{cite}
\usepackage[percent]{overpic}
\usepackage{epsfig}
\usepackage{cite}
\usepackage{hyperref}
\usepackage{color,soul}
\usepackage{multirow}
\usepackage{booktabs}
\usepackage{subfig}
\usepackage{mwe,tikz}
\usepackage[percent]{overpic}

\newcommand\copyrighttext{%
	\footnotesize \textcopyright This is the accepted version of the article, which will be published in the final form at Soft Robotics Journal. This original submission version of the article may be used for non-commercial purposes in accordance with the Mary Ann Liebert, Inc., publishers’ self-archiving terms and conditions.  
}
\newcommand\copyrightnotice{%
	\begin{tikzpicture}[remember picture,overlay]
		\node[anchor=north,yshift=-10pt] at (current page.north) {\fbox{\parbox{\dimexpr\textwidth-\fboxsep-\fboxrule\relax}{\copyrighttext}}};
	\end{tikzpicture}%
}

\begin{document}
% \title{A Soft-Bodied Aerial Robot Capable of Robust Collision and Rapid Shape Conforming, Contact Reactive, Bistable Adaptive Perching}
\title{A Soft-Bodied Aerial Robot for Collision Resilience and Contact-Reactive Perching}
\author{Pham H. Nguyen$^{+}$, Karishma Patnaik$^{+}$, \\Shatadal Mishra, Panagiotis Polygerinos and Wenlong Zhang$^{*}$% <-this % stops a space
%\thanks{*Equally Contributing 1st Authors}% <-this % stops a space
% \thanks{This work was supported in part by the National Science Foundation under Grant CMMI-1800940.}% <-this % stops a space
\thanks{K. Patnaik and W. Zhang are with the Polytechnic School, Ira A. Fulton Schools of Engineering, Arizona State University, Mesa, AZ 85212, USA. 
{\tt\small $\{$kpatnaik, wenlong.zhang$\}$@asu.edu}}%
\thanks{P. H. Nguyen was with the Polytechnic School, Arizona State University. He is currently with the Aerial Robotics Lab, Imperial College London, UK. 
{\tt\small n.pham@imperial.ac.uk}
}
\thanks{S. Mishra was with the Polytechnic School, Arizona State University. He is currently with Toyota InfoTech R\&D Labs, Mountain View, CA, USA.
{\tt\small smishr13@asu.edu}
}
\thanks{P. Polygerinos is currently with Control Systems and Robotics Laboratory (CSRL), Hellenic Mediterranean University,   Crete, Greece.
	{\tt\small polygerinos@hmu.gr}}
\thanks{$^{+}$ The first two authors contribute equally to this work.}
\thanks{$^{*}$ Address all correspondence to this author.}
}%
% \thanks{Z. Qiao and I. Mohd are with School for Engineering of Matter, Transport and Energy, Ira A. Fulton Schools of Engineering, Arizona State University, Tempe, AZ 85281, USA. {\tt\small $\{$zhi.qiao.1, imohd$\}$@asu.edu}}
% \thanks{$^+$P. H. Nguyen and K. Patnaik are equally contributing authors. $^*$Address all correspondence to this author.}}
% \thanks{P.H. Nguyen and S. Mishra were previously with the Polytechnic School,Arizona State University. P.H. Nguyen is currently with the Aerial Robotics, Imperial  College  London,  UK.  S.  Mishra  is  currently  with  Toyota  Infotech R&D, Mountain View, CA,  USA. n.pham@imperial.ac.uk, shatadal.mishra@toyota.com}}%

%% The paper headers
%\markboth{Submitted, Under Review, Soft Robotics, Mary Ann Liebert, Inc, Jan 18 2022}%


\maketitle
\copyrightnotice
%%%%%%%%%%%%%%%%%%%%%%%%%%%%%%%%%%%%%%%%%%%%%%%%%
%%%%%%%%%%%%%%%%%%%%%%%%%%%%%%%%%%%%%%%%%%%%%%%%%
%%%%%%%%%%%%%%%%%%%%%%%%%%%%%%%%%%%%%%%%%%%%%%%%%
%%%%%%%%%%%%%%%%%%%%%%%%%%%%%%%%%%%%%%%%%%%%%%%%%

\begin{abstract}

Current aerial robots demonstrate limited interaction capabilities in unstructured environments when compared with their biological counterparts. Some examples include their inability to tolerate collisions and to successfully land or perch on objects of unknown shapes, sizes, and texture. Efforts to include compliance have introduced designs that incorporate external mechanical impact protection at the cost of reduced agility and flight time due to the added weight. In this work, we propose and develop a light-weight, inflatable, soft-bodied aerial robot (SoBAR) that can pneumatically vary its body stiffness to achieve intrinsic collision resilience. Unlike the conventional rigid aerial robots, SoBAR successfully demonstrates its ability to repeatedly endure and recover from collisions in various directions, not only limited to in-plane ones. Furthermore, we exploit its capabilities to demonstrate perching where the 3D collision resilience helps in improving the perching success rates. We also augment SoBAR with a novel hybrid fabric-based, bistable (HFB) grasper that can utilize impact energies to perform contact-reactive grasping through rapid shape conforming abilities. We exhaustively study and offer insights into the collision resilience, impact absorption, and manipulation capabilities of SoBAR with the HFB grasper. Finally, we compare the performance of conventional aerial robots with the SoBAR through collision characterizations, grasping identifications, and experimental validations of collision resilience and perching in various scenarios and on differently shaped objects.
\end{abstract}


%%%%%%%%%%%%%%%%%%%%%%%%%%%%%%%%%%%%%%%%%%%%%%%%%%%%%%%%%%%%%%%%%%%%%%%%%%%%%%%%
\section{Introduction}
%%%%%%%%%%%%%%%%%%%%%%%%%%%%%%%%%%%%%%%%%%%%%%%%%%%%%%%%%%%%%%%%%%%%
%%%%%%%%%%%%%%%%%%%%%%%%%%%%%%%%%%%%%%%%%%%%%%%%%%%%%%%%%%%%%%%%%%
Aerial manipulation highlights the ability for aerial robots to perform manipulation or interaction tasks such as perching/grasping to widen their scope of applications. Perching, specifically, enhances the ability for aerial robots to save energy and maintain a vantage position for monitoring or surveillance~\cite{Danko2005,Meng_2022_perching_review} and the perching mechanism can also be exploited for performing dynamic grasping. Existing aerial robots coordinate perching mechanisms and flight dynamics in order to achieve this~\cite{Ruggiero2018AerialReview,Kovac2009AVehicles,Meng_2022_perching_review}. Nature, however, calls attention to various physically intelligent features that can enhance the proficiency of dynamic aerial robot perching and grasping~\cite{sitti2021physical,Hang2019PerchingGears,nguyen2022adopting}. Birds and bats enter a coordinated post-stall maneuver, to maintain a constant rate of approach in combination with a high angle of attack~\cite{Cory2008ExperimentsPerching, moore2012control, Pope2016PlanningSurfaces}. At impact, their feet clasp the irregular perch and their legs bend to absorb their momentum~\cite{Doyle2013AvianPercher,watson1869mechanism,chi2014Perching}. Their feet also utilize a passive tendon locking mechanism, so no additional energy is wasted during perching~\cite{Quinn1990TendonLock}. Even smaller insects, like flies, utilize a combination of collision and perching, and their compliant bodies help dampen the perching impact~\cite{Hyzer1962fly}.

%%%%%%%%%%%%%%%%%%%%%%%%%%%%%%%%%%%%%%%%%%%
%\begin{figure}
%	\centering
%	\includegraphics[width=0.48\textwidth]{Image_1_Soft_Drone_v4_smaller-01-min.png}
%	\caption{(A) Flying and dynamic collision-based perching sequence. The woven fabric utilized is highlighted under a microscope. (B) Perching mechanism sequence from straight beam to curled to recovery state. (C) Soft frame from a deflated stored state to a inflated and rigid state. (D) Head-on wall collision with SoBAR. (E) Dynamic perching to recovery to landing sequence of SoBAR. Video: \url{https://youtu.be/Xgf67ZaSvRw}}
%	\label{fig:soft_drone_fig1}
%\end{figure}
%%%%%%%%%%%%%%%%%%%%%%%%%%%%%%%%%%%%%%%%%%%

\begin{figure}
	\centering
	\includegraphics[width=0.48\textwidth]{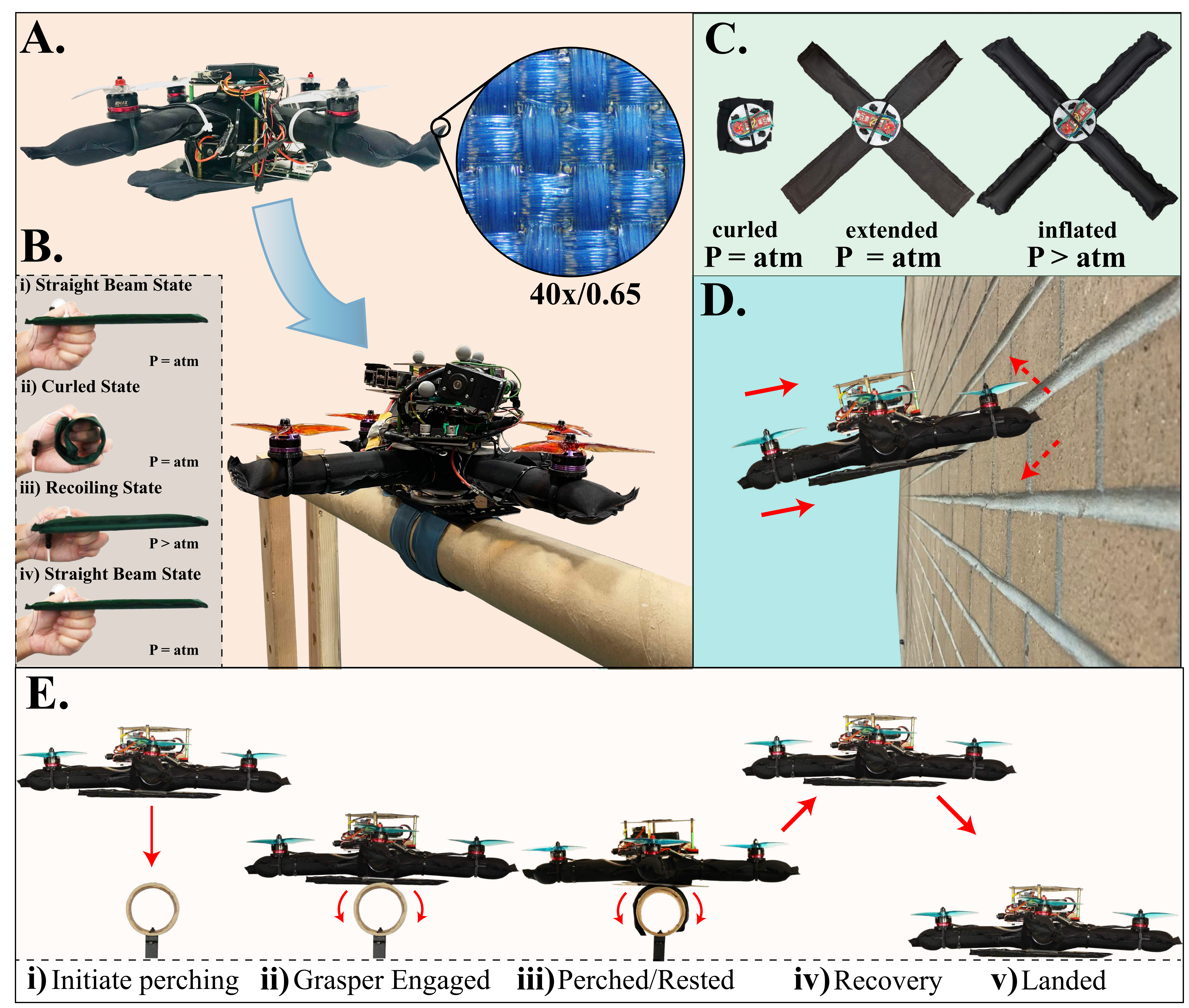}
	\caption{\textcolor{black}{The overall operational scheme of SoBAR. (A) Flying and dynamic collision-based perching sequence. The woven fabric utilized is highlighted under a microscope. The magnification factor is 40x and aperture is 0.65. (B) The hybrid fabric-based bistable (HFB) grasper operating from a straight beam (prior to perching) to curled (perching) to recovery state (pneumatic recovery after perching). (C) Soft-bodied frame from a deflated stored state to an inflated and rigid state. (D) Head-on wall collision with SoBAR. (E) Grasping sequence of actuator takes 4$\mu$s and utilizes pneumatic actuation to recoil in approximately 3s. (F) Setting up and assembling SoBAR, approximately takes 4mins. (G) Dynamic perching to recovery to landing sequence of SoBAR. Video: \url{https://youtu.be/Xgf67ZaSvRw}}}
	\label{fig:soft_drone_fig1}
\end{figure}

However, there is often a dissociation between controlled collision and dynamic perching in the existing design of aerial robots, as the rigid-body structures (found on quadrotors and avian-inspired graspers) are not good at mitigating collision impact incurred during dynamic perching. Furthermore, current avian-inspired graspers are limited to perching on cylindrical-shaped structures~\cite{erbil2013design,thomas2014toward,stewart2021passive,roderick2021bird}, this is because of the lack of intrinsic softness in their fingers and joints. Recent work has started taking into account controlled collisions during dynamic perching~\cite{stewart2021passive, roderick2021bird, kirchgeorg2021hedgehog}, utilizing mechanisms to mitigate the impact during a more dynamic perching sequence.  For example, Roderick et. al~\cite{roderick2021bird} utilizes legs that bend (along with an avian-inspired grasper with claws), while Stewart et.al~\cite{stewart2021passive} utilizes a spring-based mechanism  to absorb the robot’s momentum resulting from perching impact. Kirchgeorg et. al~\cite{kirchgeorg2021hedgehog}, on the other hand, explores the use of an external protective exoskeleton, along with a high-friction, passive, hook-and-hang perching mechanism. These robots, however, do not extensively quantify their ability to mitigate the high impact in collision-based perching. They are also still limited by  grasping targets with circular cross-sections.

Along with dynamic perching, aerial robots also have to deal with unexpected interactions in obstacle-laden environments with poor visual conditions. Therefore, collisions are inevitable even with state-of-the-art collision avoidance and computer vision systems \cite{sedaghat2017collision,spasojevic2020perception}. With aerial robots, high-energy impacts or collisions can lead to structural damage or loss of control, resulting in crashes. While conventionally, researchers have avoided physical interactions \cite{patnaik2021towards,Bucki2019DesignQuadcopter}, recently, researchers have developed collision-resilient aerial robots with compliant bodies to sustain collisions whilst remaining stable in the air and/or surviving structural damage after crashing~\cite{Mintchev2017Insect-InspiredMulticopters, dilaverouglu2020minicore, patnaik2021collision, patnaik2020design, liu2021toward} or additional external protective structures~\cite{Sareh2018RotorigamiRotorcraft, klaptocz2013euler, liu2020toward,savin2022mixed_tensodrone,mintchev2018soft,zha2020collision, briod2014collision, hedayati2020pufferbot,kirchgeorg2021hedgehog}.

Soft robotics has emerged as a promising solution to approach the problem of collision resilience and safe perching ~\cite{coyle2018,rus2015,laschi2016}. Compliant materials have been utilized to design soft or foldable wings~\cite{floreano2015science, Mintchev2016AdaptiveMorphology, di2017bioinspired,daler2015bioinspired,jafferis2019untethered}, deformable rotors~\cite{nguyen2020towards}, compliant joints and armatures~\cite{Mintchev2017Insect-InspiredMulticopters, shintake2015AntagonisticActuator, tonazzini2016variable, Sareh2018RotorigamiRotorcraft, ramezani2017biomimetic, Mintchev2018BioinspiredOrigami}, and compliant graspers or landing gears~\cite{Kim2018AnFlat, miron2018sleeved, lee2020snatcher, li158untethered, Kovac2016LearningRobots, Zhang2019ASkeletons,nguyen2019passively, soft_arms_GarciaRubiales2021,Fishman2021_Soft_Drone}. These soft solutions for perching and grasping, however, are often limited by their load bearing capabilities and slow grasping speeds. The former limits their ability to maintain a strong grasp on objects or carry meaningful payloads. Due to their limited grasping speeds, they resort to hovering closely or landing on the perch prior to grasping. They also sometimes require active actuation to maintain constant grasping or perching position, which reduces the overall system efficiency. 

% \textcolor{black}{In this work, we develop both the body and perching mechanism of an aerial robot to be able to tolerate high-impact and head-on collisions with the environment using a novel inflatable soft-bodied design.
% We further exploit the soft nature of our aerial robot to achieve collision-based perching by utilizing the generated impact forces, as seen in Figs.~\ref{fig:soft_drone_fig1}A and D. Thus, SoBAR highlights a pneumatic lightweight soft-bodied frame, capable of modulating its stiffness, for contact resilience and flight stability, as seen in Fig.~\ref{fig:soft_drone_fig1}C. The frame is developed with high-strength woven fabrics so the robot is robust to environmental interactions but lightweight and easily stowable, when deflated, as shown in Fig.~\ref{fig:soft_drone_fig1}D. The process of setting up SoBAR takes \textcolor{black}{approximately} 4 \textcolor{black}{minutes}, making it effortlessly portable, as highlighted in Fig.~\ref{fig:soft_drone_fig1}F and Supplementary Video 1.}

% We approach this task by looking at how compliant bodies of animals help dampen the perching impact~\cite{Hyzer1962fly} and how birds have air-filled cavities in their bones, essentially airframes that are lightweight yet are still strong enough, to accommodate forces during flight~\cite{sullivan2017extreme} and landing. 

In this work, we develop a novel inflatable soft-bodied aerial robot (SoBAR) that can tolerate high-impact collisions with the environment in any direction as seen in Figs.~\ref{fig:soft_drone_fig1}A and D. SoBAR highlights a  light-weight pneumatic frame capable of modulating its stiffness, for contact resilience and flight stability, as seen in Fig.~\ref{fig:soft_drone_fig1}C. The process of setting up SoBAR takes \textcolor{black}{approximately} 4\textcolor{black}{min}, making it effortlessly portable, as highlighted in Fig.~\ref{fig:soft_drone_fig1}F and Supplementary Video 1.
	% \par
	% In \textcolor{black}{tandem to the} collision-mitigation abilities
	Further capitalizing on the collision resilient frame, we introduce a passive dynamic hybrid fabric-based bistable (HFB) grasper \textcolor{black}{design, which} reacts to impact upon contact with the perching surface and enables manipulation abilities, \textcolor{black}{as illustrated} in Fig.~\ref{fig:soft_drone_fig1}B and E. \textcolor{black}{Utilizing a snap-through buckling instability in its passive design,} the grasper \textcolor{black}{absorbs} the impact energy \textcolor{black}{and uses it to}  \textcolor{black}{transform} into a continuum \textcolor{black}{closed-form} grasping shape \textcolor{black}{in about 4ms, to adapt to various shapes and sizes of perching objects}. An additional advantage of such design approach is that the passive grasper does not utilize any additional energy to maintain grasp, and can be pneumatically retracted in less than 3s, as seen in Fig.~\ref{fig:soft_drone_fig1}E. 
Together, the soft-bodied frame and HFB grasper, operate synergistically by mitigating the impact for successful dynamic perching  on various objects as seen in Supplementary Video S5 and S6 and Fig.~\ref{fig:soft_drone_fig1}G . 

% Together, the soft-bodied frame and HFB grasper, operate synergistically. Through their embodied physically intelligent features
% \cite{sitti2021physical,miriyev2020skills} the system's perching performance is maximized. This is achieved by mitigating the impact during dynamic perching to extend the time for grasping, and adapting to the shape and size of the object to perch, as seen in Supplementary Video S5 and S6. Finally, we successfully demonstrate the ability of SoBAR to autonomously perch and recover on various sized and shaped objects, as seen in Fig.~\ref{fig:soft_drone_fig1}G. 

\textcolor{black}{To the best of our knowledge, this is the first report of a multirotor aerial robot that utilizes an entirely soft body based on inflatable fabrics to modulate its stiffness and  absorb impacts.} In summary, this work contributes to:

\textcolor{black}{
	\begin{enumerate}
		\item  A new class of aerial robot frames based on inflatable high-strength fabrics, capable of effectively absorbing impact forces from collisions and contact-based perching.
		\item A new class of a lightweight gripper for aerial robots based on the combination of inflatable high-strength fabrics and a bistable mechanism, which is capable of providing passively activated dynamic perching on objects of unknown shape and size.
		\item Modeling and analysis of both, the developed soft aerial robot and the bistable grasper, for developing planning and control strategies.
		% 	\item  A planning and control strategy for the proposed compliant aerial robot (SoBAR) to successfully approach and perch utilizing the compliant grasping mechanism (HFB). 
	\end{enumerate}
}

%%%%%%%%%%%%%%%%%%%%%%%%%%%%%%%%%%%%%%%%%%%%%%%%%%%%%%%%%%%%%%%%%%%%%%%%%%%%%%%%%%%%%%%%%%%%%%%%%%%%%%%%%%%%%%%%%%%%%%%%%%%%%%%%%%%%%%%%%%%%%%%%%%%%%%%%%%%%%%%%%%%%%%%%%%%%%%%%%%%%%%%%%%%%%%%%%%%%%%%%%%%%%%%%%%%%%%%%%%%%%%%%%%%%%%%%%%%%%%%%%%%%%%%%%%%%%%%%%%%%%%%%%%%%%%%%%%%%%%%%%%%%%

\section{Materials and Methods}
\label{sec:concept}
\begin{figure*}[t]
	\centering
	\includegraphics[width =0.85\textwidth]{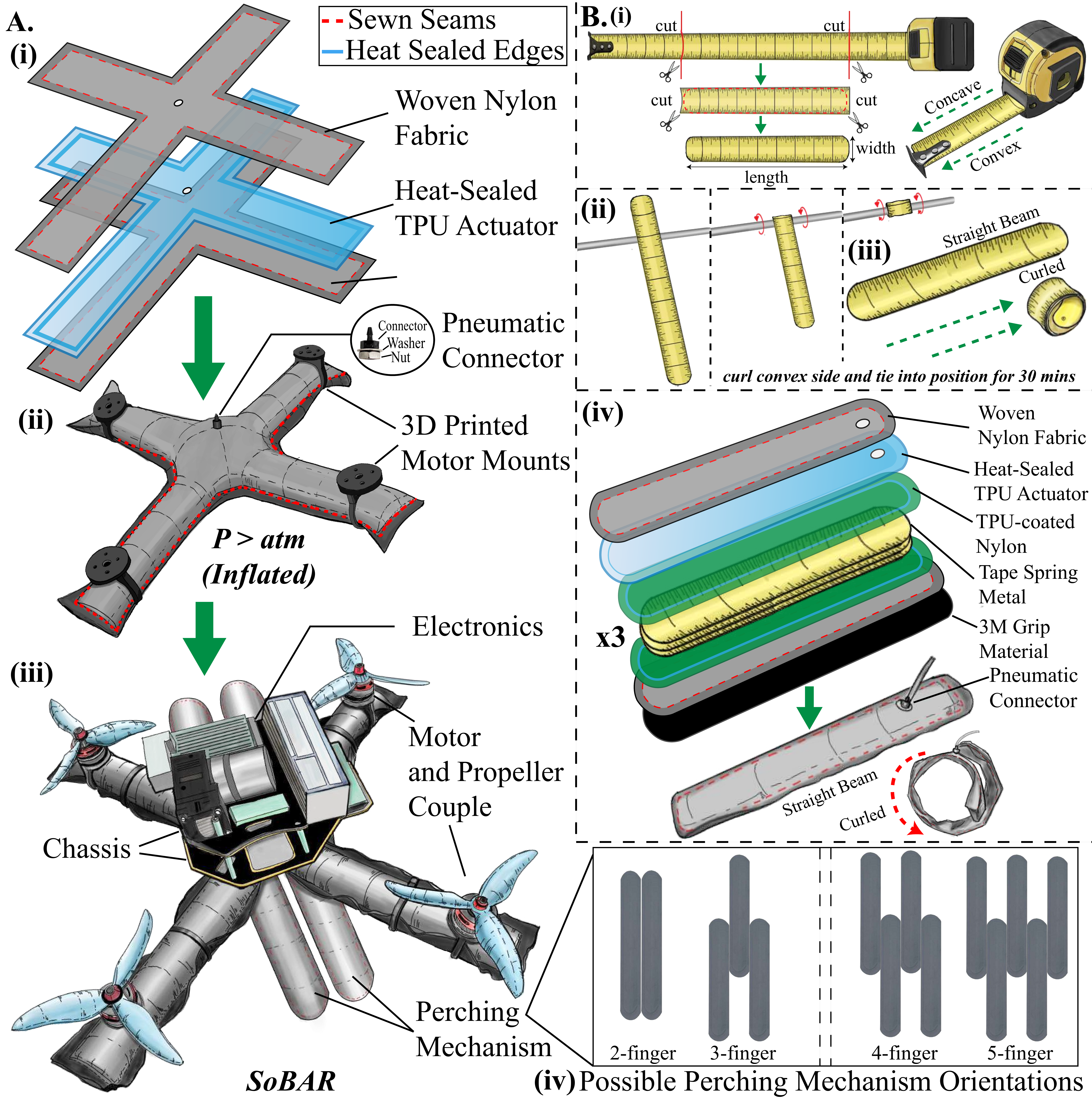}
	\caption{\textcolor{black}{The manufacturing process of SoBAR and the HFB actuators.} (A) Fabrication steps for SoBAR.  (i) First, align the woven fabric and heat-sealed actuator, after laser cutting. The woven fabric sheets are then sewn along the edges, utilizing a super-imposed seam, and the he heat-sealed actuator is inserted, to create the soft frame. (ii) Next, the pneumatic connector and 3D-printed motor mounts are added and aligned on the inflated soft-bodied frame. (iii) Finally, mount the propellers and motor pairs, electronics, and the perching mechanism. (iv) Possible perching mechanism orientations made of multiple HFB actuators (B) HFB actuator fabrication. (i) First, prepare the curling bistable mechanism by cutting and forming the spring steel metal. (ii) Next, curl the bistable tape spring, along its convex side, and  \textcolor{black}{restraining} this position for \textcolor{black}{a minimum of 30 min}. (iii) The curling bistable tape spring. (iv) Finally, align the woven fabric, heat-sealed TPU actuator, TPU-coated nylon, three tape springs, and 3M grip material to create a single HFB actuator.}
	% \subfloat[Wrench Hull Obtained]{\includegraphics[width = 0.4\textwidth]{}}
	\label{fig:fabrication}
	\vspace{-1em}
\end{figure*}
%%%%%%%%%%%%%%%%%%%%%%%%%%%%%%%%%%%%%%%%%%%%%%%%%%%%%%%%%%%
%%%%%%%%%%%%%%%%%%%%%%%%%%%%%%%%%%%%%%%%%%%%%%%%%%%%%%%%%%%

\subsection{Design of Soft-Bodied Structure}
\label{sec:soft_body}

We opted for a standard ``$\times$'’ or ``$+$’' configuration for designing SoBAR’s pneumatic frame, to \textcolor{black}{benchmark} the system’s mechanical resilience in one arm or two arm collisions. We designed SoBAR’s frame to be geometrically similar to DJI F450’s standard rigid frame (319mm$\times$319mm), for comparison in collision tests. SoBAR’s and DJI’s frames weigh 10g and 120g, respectively.  

To develop the soft-bodied frame, we took inspiration from the light-weight, thin-walled, hollow pneumatic bones that are found in bird wings~\cite{sullivan2017extreme,novitskaya2017reinforcementsbones}. 
% These pneumatic bones are connected to their pulmonary system that regulates air to increase skeletal buoyancy, making it extremely lightweight but still very resistant against external loads~\cite{novitskaya2017reinforcementsbones}. 
Combining this idea with our previous work \textcolor{black}{on soft fabric-based actuators}~\cite{nguyen2020design,sridar2017development,sridar2018soft}, we chose to evoke a soft-bodied structure based on thin-walled, soft inflatable fabric beams\textcolor{black}{. Such pneumatically driven structures are} shown to be mechanically resilient to external interactions and \textcolor{black}{can absorb} impact-induced energies. These characteristics enable SoBAR to handle high-speed collisions, collision-based perching, and emergency landings. Additionally, by having a collision-safe air frame, \textcolor{black}{eliminates the} need for a cage-like structure around the aerial robot \textcolor{black}{in applications where no humans are present}, thus making the design compact and efficient. The intrinsically soft platform we designed can also vary in stiffness through pneumatic activation. Fig.~\ref{fig:soft_drone_fig1}C and F shows that at zero internal pressure, the frame is completely collapsible and each arm can compress from 20.5cm to 3cm\textcolor{black}{; a reduction in length of} \textcolor{black}{85\%} (also shown in Supplementary Video 1). For flight, the frame is inflated to its maximum stiffness to reduce undesired oscillations, instabilities, or slow \textcolor{black}{flight maneuver} responses. The SoBAR frame can absorb impact through deformation, which extends the impact time with the perching objects to support the collision-based passive perching maneuver with the HFB grasper, as seen in Fig.~\ref{fig:soft_drone_fig1}A. Fabrication of the SoBAR's frame is detailed in Fig.~\ref{fig:fabrication}A and Appendix~A.

%%%%%%%%%%%%%%%%%%%%%%%%%%%%%%%%%%%%%%%%%%%%%%%%%%%%%%%%%%%%%
\subsection{Design of the HFB Grasper}
The HFB perching structure is designed with a TPU-coated nylon fabric external structure encasing pre-formed bistable metal steels, capable of converting high-impact energy and instantly reacting to the contact, in order to go from a straight beam to a rapidly curling state, as seen in Fig.~\ref{fig:soft_drone_fig1}B. Each perching structure is made of multiple HFB actuators placed in possible perching orientations, as in in Fig.~\ref{fig:fabrication}B and Fig.~\ref{fig:fabrication}A(iv). 

%The foundation of this design stems from the ability to create structures with controlled buckling. 

The design combines the energy storage nature of deformable spring steels and fabric-based actuators. The bistable spring steel, when activated, leads to power amplification and rapid curling movements that are highly desired for grasping. Furthermore, after perching, no further mechanical activation is required. In order to release it from its perched state, the inflatable fabric-based actuator layer allows the system to quickly recoil, to its initial straight beam state, as in Fig.~\ref{fig:soft_drone_fig1}B(iv) and Supplementary Video 2. 
% After which, no further activation is required to maintain this initial state, . 
In this state, the grasper can also be utilized as landing skids. Along with its soft-bodied frame, the robot is capable of safe emergency landing situations, as seen in Fig.~\ref{fig:soft_drone_fig1}G(v). Fabrication of the HFB grasper is detailed in Appendix~A.

%Here, we utilized three spring steels to improve the grasping force of the actuator. These orientations can also be developed utilizing the same fabrication method. 

%%%%%%%%%%%%%%%%%%%%%%%%%%%%%%%%%%%%%%%%%%%%%%%%%%%%%%%%%%%
\subsection{Hardware Overview}
%%%%%%%%%%%%%%%%%%%%%%%%%%%%%%%%%%%%%%%%%%%%%%%%%%%%%%%%%%%

\begin{figure}[t]
	\centering
	\includegraphics[trim = 5cm 2cm 5cm 2cm, clip, width=0.48\textwidth]{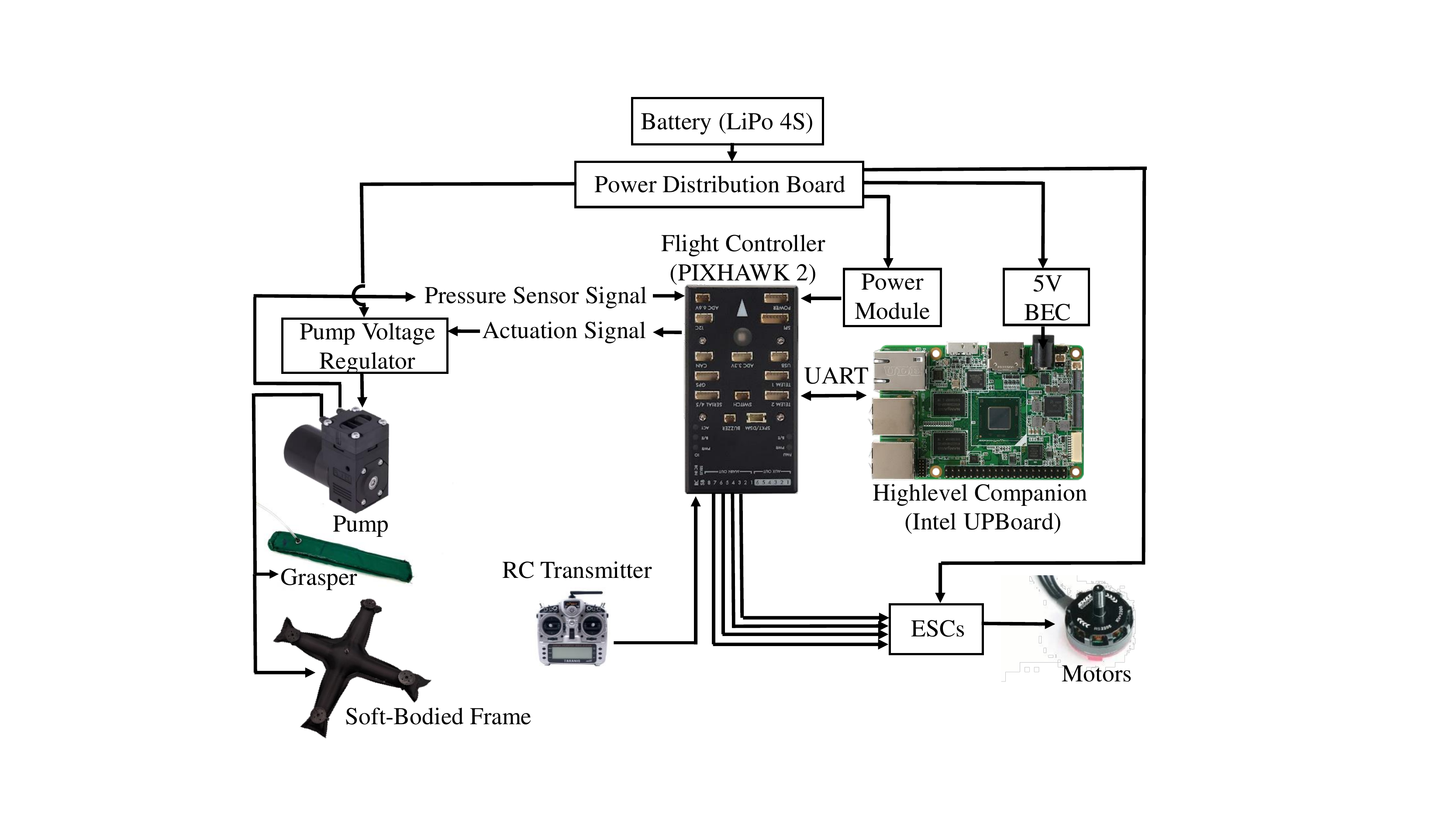} %maroon
	\caption{Electronics diagram of SoBAR. \textcolor{black}{On the right part of the diagram all the necessary components to achieve flight are illustrated. On the left part of the diagram the proposed electro-pneumatic components for the soft frame and perching grasper are shown.}}
	\label{fig:electronics}
	\vspace{-1em}
	% \vspace{-1.5em}
\end{figure}
%%%%%%%%%%%%%%%%%%%%%%%%%%%%%%%%%%%%%%%%%%%%%%%%%%%%%%%%%%%

SoBAR's chassis hosts the flight controller, power module, and high-level companion computer, as seen in Fig.~\ref{fig:electronics}. The flight controller monitors and controls the internal pressure of SoBAR’s body and the HFB grasper, by utilizing analog pressure sensors and an on-board micro diaphragm pump. The specific hardware and electronics details are further documented in Appendix~B. Prior to flight, motor propeller pairs are aligned and the flight controller sensors with the air frame are calibrated through QGroundControl. The experimental setups utilized in this work, including the universal tensile testing machine, the high-speed camera, high-G accelerometer, and motion capture system are detailed in Appendix~C.

\begin{figure*}[t]
	\centering
	\includegraphics[trim = 1cm 2cm 0cm 0cm, clip, width = \textwidth]{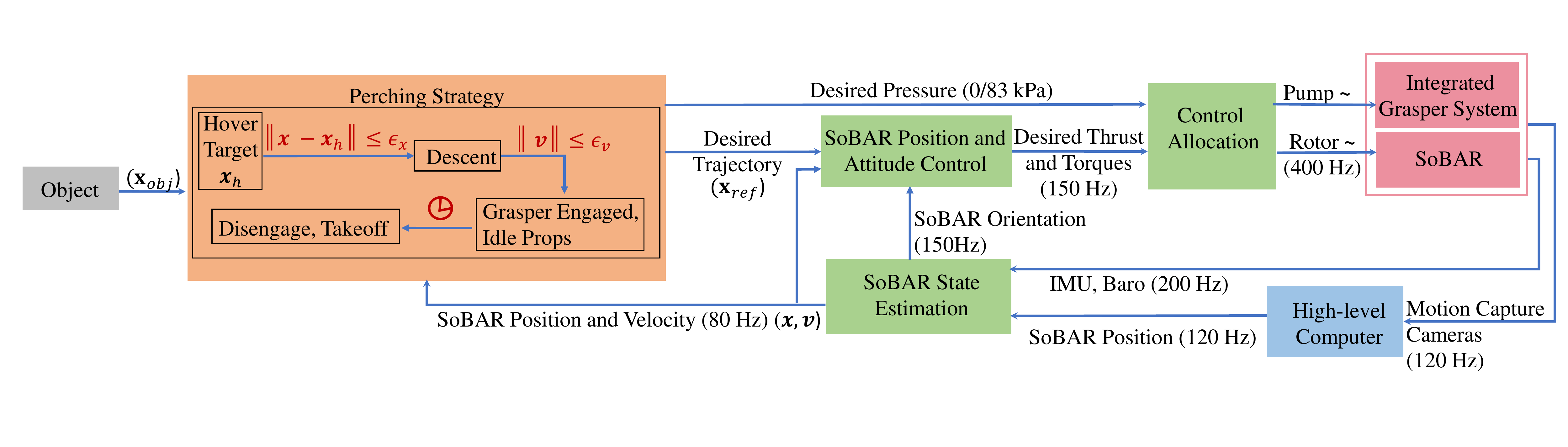}
	\caption{The complete closed-loop control pipeline of SoBAR for the perching task. The green blocks show the computation performed on the flight controller. The high-level companion computer is used to relay the position and orientation information of SoBAR from the indoor positioning system to the flight controller. The perching strategy, shown in the orange block, represents the state machine during an autonomous perching task. Mathematical conditions represent event-triggered transitions while the clock symbol represents the time-triggered ones. Here, $x_h$ refers to the hover target location for SoBAR directly above the perching target before initiating the descent. After the errors in position are within a tolerance region denoted by $\epsilon_x$, SoBAR initiates the descent trajectory. Once the grasper is engaged, the velocities are almost zero to indicate that the SoBAR has perched. After a user defined wait time, it then performs recovery control by first disengaging the grasper and then taking-off. }
	\label{fig:cbd}
	\vspace{-1em}
\end{figure*}
%%%%%%%%%%%%%%%%%%%%%%%%%%%%%%%%%%%%%%%%%%%%%%%% 
\subsection{Modeling and Control of SoBAR}
\label{sec:modelling_control}

%%%%%%%%%%%%%%%%%%%%%%%%%%%%%%%%%%%%%%%%%%%%%%%%
\begin{figure}[b]
	\centering
	\includegraphics[width=0.25\textwidth]{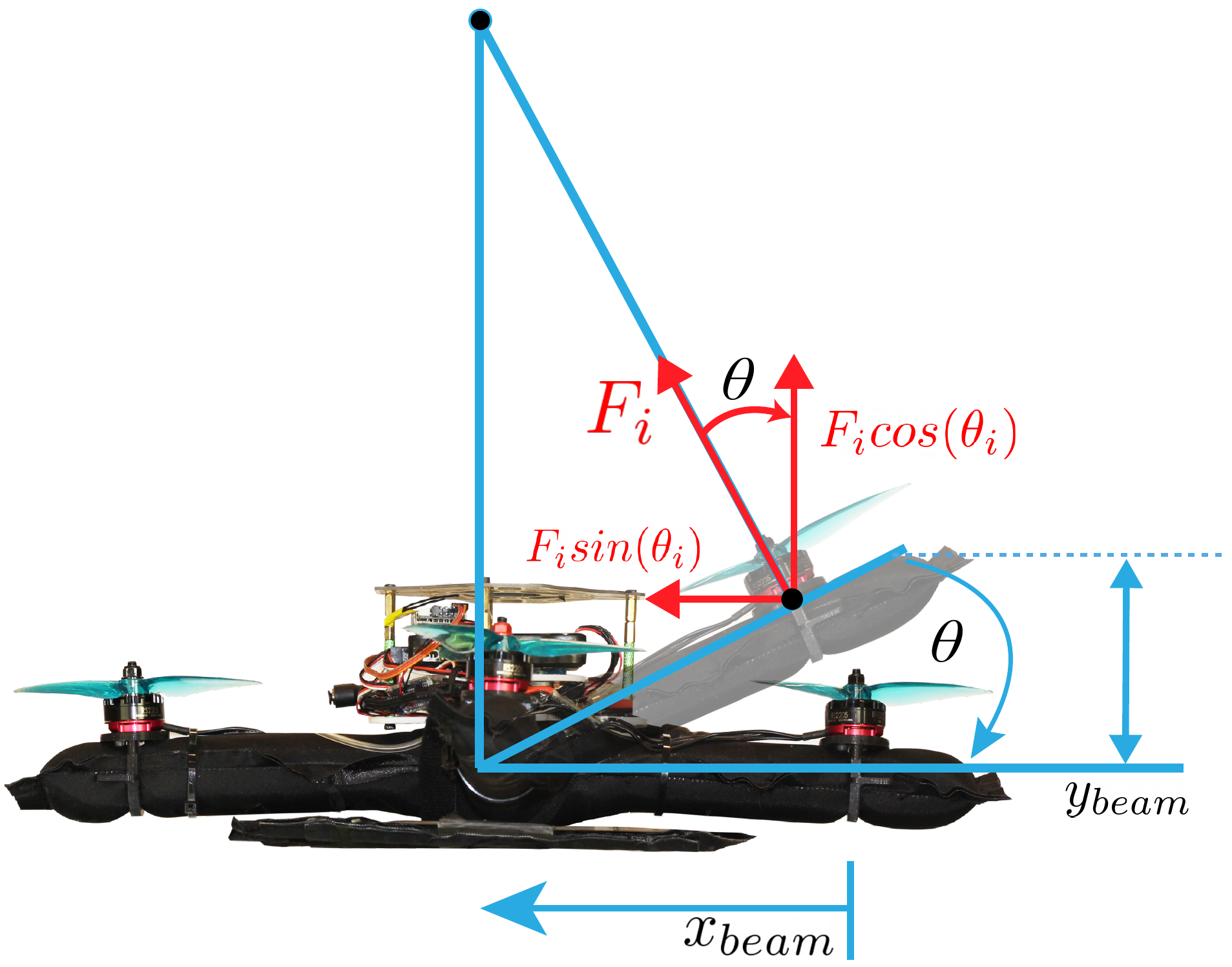}
	\caption{Modeling arm deflection ($\theta_i$) versus internal pressure to estimate the thrust loss coefficient due to arm bending}
	\label{fig:arm_bending}
	% \vspace{-1em}
	% \vspace{-1.5em}
\end{figure}
%%%%%%%%%%%%%%%%%%%%%%%%%%%%%%%%%%%%%%%%%%%%%%%%

As shown in Fig. \ref{fig:arm_bending}, the rotor thrust is significant enough to introduce slight deflection of the soft arms affecting in this way net thrust and SoBAR's flight performance. To address this issue, we derive a model for the thrust loss coefficient as a function of the arm deflection angle.

As noted from our experiments, discussed in detail in Sec. 3.2.1, the maximum beam deflection at 207kPa was noted to be less than 10$^o$. Also for inflatable beams without wrinkles, we assume that the plane sections remain plane, i.e there are no significant shear or torsional stresses relative to the bending (axial) stresses. Based on these assumptions,  we employ the Euler–Bernoulli beam theory \cite{sanan2009robots,howell2013compliant} to model the beam deflection as follows:

\begin{equation}\label{eqn:beam_eqn}
	\centering
	\begin{aligned}
		y_{beam} &= \frac{F_il^3}{3EI}, \\
		\theta_i &= - \frac{F_il^2}{2EI},
	\end{aligned}
\end{equation}
where $l$, $F_i$, and $I$ are the arm length, thrust produced by the $i^{th}$ motor, and second moment of area of the beam, respectively. $y_{beam}$ is the experimentally measured tip deflection as shown in Fig. \ref{fig:arm_bending}.  

Denoting the inertial reference frame by $\{i_1, i_2, i_3\}$ and the body fixed frame by $\{b_1, b_2, b_3\}$, the deflection angle, $\theta_i$, can then be used to estimate the net effective thrust by (\ref{eqn:theta_def}) as shown in Fig. \ref{fig:arm_bending}.
\begin{equation}\label{eqn:theta_def}
	f_i = F_i \cos \theta_i,
\end{equation}
where $f_i$ denotes the effective thrust of the $i^{th}$ propeller along the $-b_3$ axis and the thrust loss coefficient is approximated by $\cos \theta_i$. For the controller design, we consider the controller inputs as the total thrust and moments on the system, $\textbf{u} = [f~\textbf{\textit{M}}]^T$, which are related to the effective $i^{th}$ motor thrust according to the following allocation matrix, $A$:
% \begin{equation}
% \centering
%     \begin{aligned}
%         f &= \sum_{i=1}^4 f_i \\
%         M &= [d(f_4 - f_2) ~d(f_1 - f_3) ~ c_{\tau f} (-f_1 + f_2 -f_3 + f_4) ]
%     \end{aligned}
% \end{equation}
\begin{equation}\label{eqn:cam}
	\centering
	% \begin{pmatrix}
	%     f \\
	%     M_1 \\
	%     M_2 \\
	%     M_3 \\
	%     \end{pmatrix} = 
	\begin{pmatrix}
		f \\
		\textbf{\textit{M}}
	\end{pmatrix} =
	\underbrace{\begin{pmatrix}
			1 & 1 & 1 & 1 \\
			0 & -d &0 &d \\
			d &0 &-d &0 \\
			-  c_{\tau f} & c_{\tau f} & - c_{\tau f} & c_{\tau f}\\
	\end{pmatrix}}_\text{$A$} 
	\underbrace{\begin{pmatrix}
			f_1\\
			f_2 \\
			f_3 \\
			f_4 \\
	\end{pmatrix}}_\textbf{\textit{N}}, 
\end{equation} 
where $d$ and $c_{\tau f}$ are the distance between vehicle COM to motor and coefficient for reaction torque, respectively. 
The individual motor thrust force $(F_i)$ and the corresponding rotor speeds can then be calculated using (\ref{eqn:theta_def}) and (\ref{eqn:cam}).

The rigid body equations for SoBAR are given by:
	\begin{subequations}
		\begin{align}
			% \dot{\textbf{\textit{x}}} &= \textbf{\textit{v}} \\
			\dot{\boldsymbol{x}} &= \boldsymbol{v} \label{eqn:4a} \\
			m\dot{\boldsymbol{v}} &= mg{\textbf{\textit{e}}}_3 - f {\textbf{\textit{R}}}{\textbf{\textit{e}}}_3 + \Delta_f \label{eqn:4b} \\
			\dot{\textbf{\textit{R}}} &= \textbf{\textit{R}}\hat{\boldsymbol{\Omega}}  \label{eqn:4c} \\
			\textbf{\textit{J}}\dot{\boldsymbol{\Omega}} &= \textbf{\textit{M}} - \boldsymbol{\Omega} \times \textbf{\textit{J}}\boldsymbol{\Omega} + \Delta_\tau \label{eqn:4d}
		\end{align}
\end{subequations}
where 
% 	$\textbf{x}~=~ [\boldsymbol{x} ~ \boldsymbol{v}]^T$, 
the hat map transforms a vector into its equivalent matrix, to represent cross product as a matrix multiplication. Equation (\ref{eqn:4a}) and (\ref{eqn:4b}) describe the translational dynamics while (\ref{eqn:4c}) and (\ref{eqn:4d}) are used to describe the rotational dynamics of any quadrotor system and are standard nonlinear quadrotor dynamic equations in the SE(3) space \cite{lee2010geometric}.

A P-PID structure for the low-level position control loop with a geometric controller \cite{lee2010geometric} for the attitude control loop is employed for SoBAR's tracking control. In free flight and near-hover condition (when there is no external physical force acting on the drone), the beam deflection gives rise to drift forces in the horizontal plane due to the residuals of the motor thrust's horizontal components. The gains of the controller are tuned via multiple experimental trials such that the pitch and roll gains are able to overcome the drift forces and maintain a stable hover condition, as documented in Appendix K.   

%%%%%%%%%%%%%%%%%%%%%%%%%%%%%%%%%%%%%%%%%%%%%%%%%%%%%%%%
%%%%%%%%%%%%%%%%%%%%%%%%%%%%%%%%%%%%%%%%%%%%%%%%%%%%%%%%%%%%%%%%%%%%%%

%%%%%%%%%%%%%%%%%%%%%%%%%%%%%%%%%%%%%%%%%%%%%%%%%%%%%%%%%%%%%
%%%%%%%%%%%%%%%%%%%%%%%%%%%%%%%%%%%%%%%%%%%%%%%%%%%%%%%%%%

\subsection{Trajectory Planning for Autonomous Perching}\label{sec:planning_perch}
% This section describes the trajectory planning for an autonomous preching task with SoBAR and the HFB grasper. 
The perching maneuver consists of multiple trajectories which are described in this section. First, SoBAR approaches the target location and localizes itself to hover 30cm above it (more details are given in Appendix~D. Once the position error is within the pre-defined tolerance, SoBAR initiates the descent trajectory. The reference trajectory for the descent consists of only position setpoints. When the position errors with respect to the target perch location is approximately zero, the grasper hits the target and activates to initiate the high-impact dynamic perching. After this event, as soon as the position and velocity errors are reduced below a user defined tolerance, the propellers are turned off to remain perched. In this paper, a manual recovery is then performed by pneumatically disengaging the grasper before the take-off and landing sequence. A block diagram for autonomous perching with SoBAR is given in Fig. \ref{fig:cbd} with the perching strategy highlighted in the orange sub-block.
%%%%%%%%%%%%%%%%%%%%%%%%%%%%%%%%%%%%%%%%%%%%%%%%%%%%%%%%%%%%%%%%%%%%%%%%%%%%
%%%%%%%%%%%%%%%%%%%%%%%%%%%%%%%%%%%%%%%%%%%%%%%%%%%%%%%%%%%%%%%%%%%%%%%%%%%%%
\section{Results}\label{sec:results}
\subsection{Grasper Evaluation}
\label{sec:grasper_eval}

To characterize the performance of the HFB grasper, we first evaluated each HFB actuator for its tip force, activation force, and activation and recoil time. These experiments are detailed in Appendix~E. From the tests, we decided to utilize a triple spring steel set, that generates a grasping force of 200N, tip force of 0.55N, activates within 4ms, and pneumatically recoils within 3s with a minimum input pressure of 83kPa. From the activation force tests, we were able to approximate its desired impact velocity. With an impact time of approximately 0.1s (captured by the 500-fps high-speed camera), the triple spring steel set leads to a minimum approach velocity of 2.4m/s, which corresponds to a free-fall drop height of approximately 30cm. This insight is effectively employed to demonstrate successful perching as shown in Sec. \ref{sec:perch_eval}.

% evaluated the multi-fingered grasper. To test the grasping force of the multi-finger actuators for perching, we horizontal grasping force with the UTM. 

%%%%%%%%%%%%%%%%%%%%%%%%%%%%%%%%%%%%%%
\begin{figure}[t]
	\centering
	\includegraphics[width=0.48\textwidth]{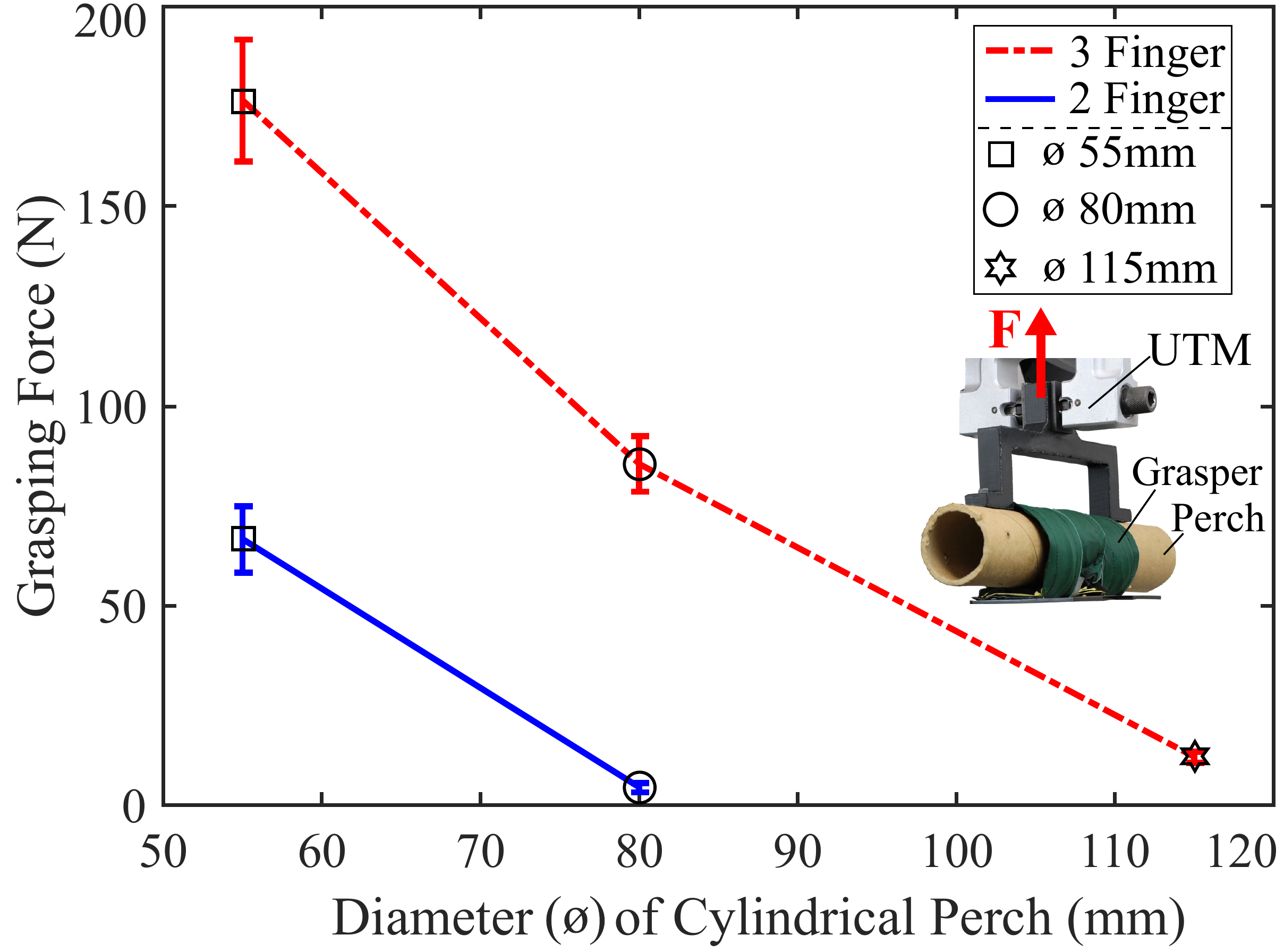}
	\caption{Grasping force characterization of different grasper configurations (two and three fingered graspers) and three perch diameters. The two-fingered grasper was not able to successfully grasp the largest \o{} = 115mm perch.}
	\label{fig:grasping_force}
	\vspace{-1em}
	% \vspace{-1.5em}
\end{figure}

We then designed the experiments with UTM for evaluating the grasping force of the multi-fingered grasper for perching. We tested the maximum grasping force for the two- and three-fingered actuator configurations, as seen in Fig.~\ref{fig:grasping_force}. For each configuration, experiments were conducted on three cylinder diameters (55mm, 80mm and 115mm), chosen based on various perch sizes of interests. The graspers were fixed in place, in a horizontal position, while the cylinders were pulled upwards at a rate of 8mm/s. As soon as slip was detected, the grasping force was recorded, as shown in Fig.~\ref{fig:grasping_force}. The average value of the maximum grasping force is calculated over five trials for each configuration. Further details of the slip detection criteria are described in Appendix L.

For the two-fingered grasper, the grasping capacity is observed as 66.58$\pm$7.39N and 4.44$\pm$1.02N for the 55mm and 80mm diameter cylinders, respectively. For the three-fingered grasper, the grasping force on the 55mm, 80mm and 115mm diameter cylinders is 176.43$\pm$12.46N, 85.4$\pm$5.55N, and 12.06$\pm$1.53N, respectively. We notice that both grasper configurations struggle to maintain grasp with the 115mm diameter cylinder because they are not able to maintain an envelope grasp around it. 

In order to study the grasping force for objects that do not conform to the HFB graspers workspace, we perform a static wrench analysis detailed in the Appendix~F. With objects within its grasp radius, the grasper has a higher chance of resisting the external wrench in order to perform a successful grasp. 

We finally demonstrate the improved grasping ability of the HFB grasper by comparing it's high power-to-weight ratio of 600 N/kg and 1173 N/kg, for the two-fingered and three-fingered version, respectively, with the many grippers available in literature\cite{Meng_2022_perching_review} as described in Appendix~I and Supplementary Table 3. 

%%%%%%%%%%%%%%%%%%%%%%%%%%%%%%%%%%%%%%%%%%%%%%%%%%%%%%%%%%
\subsection{Evaluation of Soft-bodied Frame}
\begin{figure}[b]
	\centering
	\includegraphics[width=0.48\textwidth]{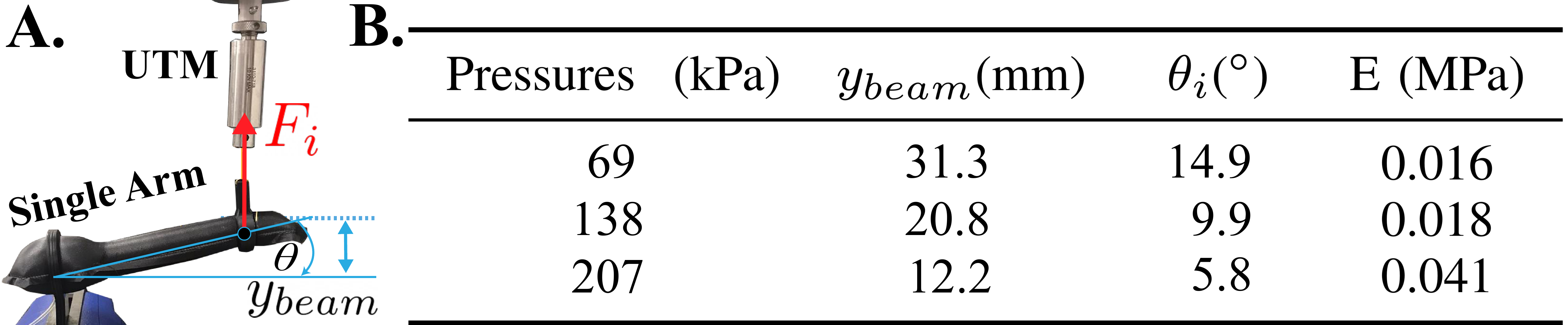}
	\caption{(A) The UTM test done to calculate the bending moment acting on a single arm of SoBAR. (B) Table for deflection characterization at different internal pressures}
	\label{fig:bending_moment}
	% \vspace{-1em}
	% \vspace{-1.5em}
\end{figure}
\subsubsection{Maximum Bending Deflection}\label{sec:beam_bend}
%%%%%%%%%%%%%%%%%%%%%%%%%

%%%%%%%%%%%%%%%%%%%%%%%%%%%%%%%%%%%%%%%%%%%%%%
A bending test was performed to calculate the maximum beam deflection due to the motor thrust at different internal pressures. The UTM was used to simulate the motor thrust as shown in Fig.~\ref{fig:bending_moment}A. With the chosen motor-propeller pair, a maximum thrust of 10N was generated by each motor and hence, the UTM was programmed to pull one end of the beam until it reached 10N. The deflection at 10N was averaged across 10 trials and denoted in Fig. \ref{fig:bending_moment}A. Approximating the arm as a circular cross-section for the second moment of area such that $I = \frac{1}{4}\pi r^4$, and with $F_i$ = 10N, $l= 0.18$m, $r$ = 15mm, we employed (\ref{eqn:beam_eqn}) and (\ref{eqn:theta_def}) to calculate the Modulus of Elasticity ($E$) and corresponding tip deflection angle ($\theta_i$) for various internal pressures as summarized in the table of Fig. \ref{fig:bending_moment}B. We see the least deflection at $y_{beam}$ = 12mm, which corresponds to a deflection angle of $\theta_i$ = 5.8$^o$ at 207kPa and the largest deflection at 69kPa with $\theta_i$ = 14.93$^o$.
%%%%%%%%%%%%%%%%%%%%%%%%%%%%%%%%%%%%%%%%%%%%%%%%%%%%%%%%%%%

%%%%%%%%%%%%%%%%%%%%%%%%%%%%%%%%%%%%%%%%%%%%%%%%%%%%%%%%%%%%%

%%%%%%%%%%%%%%%%%%%%%%%%%%%%%%%%%%%%%%%%%%%%%%%%%%%%%%%%%%%%%

\begin{figure*}[t!]
	\centering
	\includegraphics[width =0.90
	\textwidth]{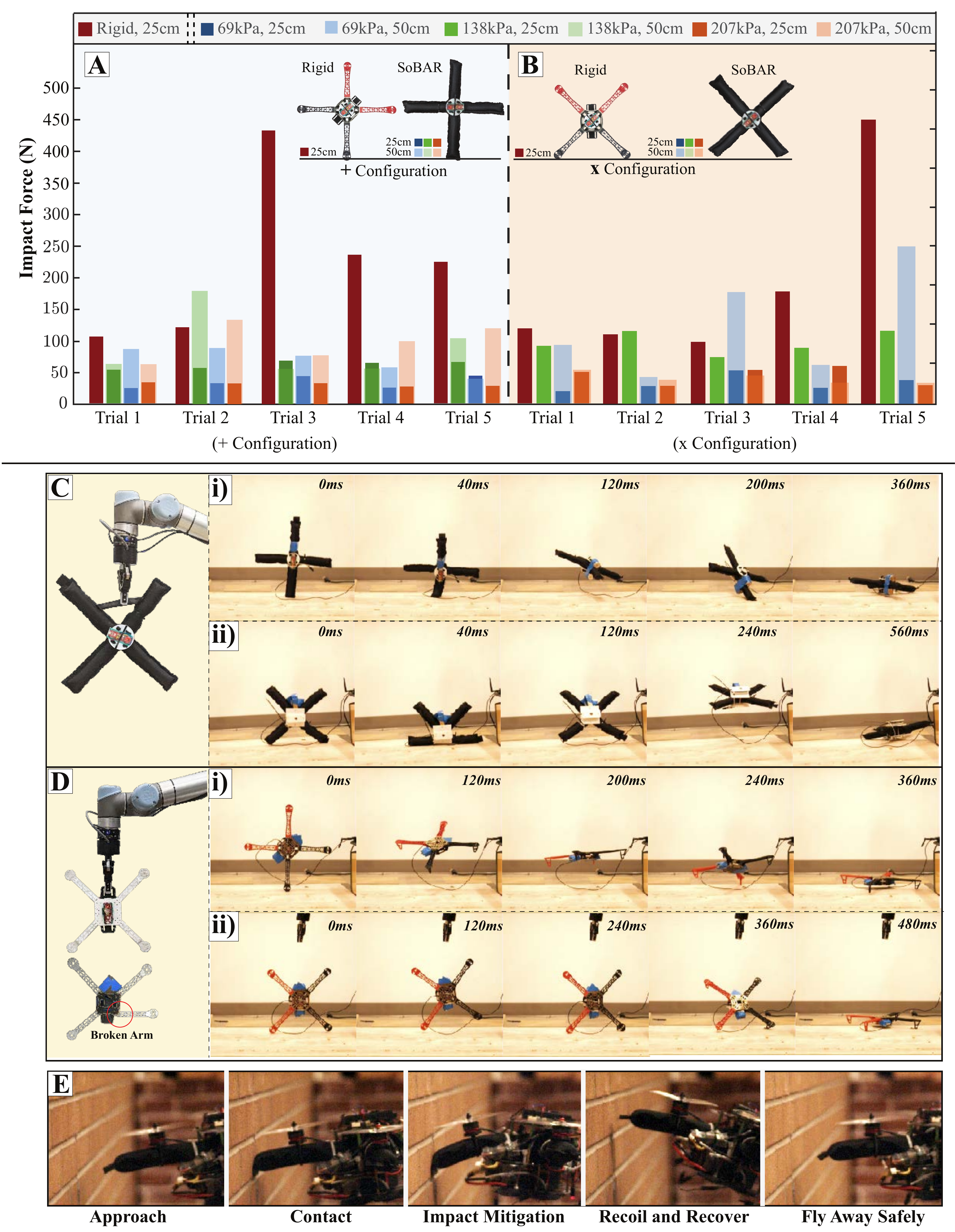}
	\caption{Collision drop tests at drop heights 25cm and 50cm for (A) rigid DJI F450 and SoBAR Frame in `+' configuration and (B) rigid DJI F450 and SoBAR in `x' configuration, were carried out for five trials each. For each trial, the left most column denotes the drop test for the rigid frame while the rest three columns denote those for the SoBAR frame with two different drop heights overlaid. Note that the impact force at 69kPa, 25cm drop height is omitted from the graph as the internal pressure of the soft frame  cannot provide cushioning to protect the rigid components of the SoBAR. By increasing the internal pressure of the soft frame the impact force can be dramatically minimized. The rigid frame was not tested for 50cm drop height since it broke on the first test as seen in (D). (C) The collision drop test with UR5 robot arm for the soft-bodied frame (i) in `+' and (ii) in `x' configuration, with internal pressures of 138kPa at 50cm drop height. (D) The collision drop test with UR5 robot arm for the rigid DJI F450 frame (i) in `+' and (ii) in `x' configuration, at 25cm drop height. Note that the frame broke due to the test at a drop height of 50cm. (E) Wall collision demonstration.}
	% \subfloat[Wrench Hull Obtained]{\includegraphics[width = 0.4\textwidth]{}}
	\label{fig:boxplot}
	\vspace{-1em}
	
\end{figure*}
%%%%%%%%%%%%%%%%%%%%%%%%%%%%%%%%%%%%%%%%%%%%%%%%%%%%%%%%%%%%%

\subsubsection{Collision Drop Tests} 
To test the collision resilience of SoBAR’s soft-bodied frame, we performed comparative drop tests with a rigid DJI F450 frame and recorded the impact times and peak accelerations of the frames in ``+'' and ``$\times$'' configurations. We also evaluated the drops at two different set heights of 25cm and 50cm corresponding to two different impact velocities of 2.21m/s and 3.1m/s respectively. For each configuration, five trials were performed and the impact times captured with a high-speed camera, shown in Supplementary Video 2. The results are shown as the bar plots in Fig.~\ref{fig:boxplot}A and B, and summarized in Table~\ref{tab:collision}. Figs.~\ref{fig:boxplot}C and D, represent the experimental setup utilized and is further detailed in Appendix~C. Slow motion frame captures of a single drop test for the ``+'' and `$\times$' configurations of the soft-bodied frame at 138kPa, are shown in Figs.~\ref{fig:boxplot}C(i) and (ii) \textcolor{black}{and also compared side by side in Supplementary Video 3}. It is also similarly displayed for the rigid frame in Figs.~\ref{fig:boxplot}D(i) and (ii). The rest of the drop tests are presented in \textcolor{black}{Appendix~H}.

For the rigid frame, the impact time for the ``$\times$'' and ``+'' configurations was approximately 22ms and 8ms. The maximum peak acceleration experienced by the frame was approximately 390m/s$^2$, corresponding to a very high peak impact force of 430N, in the ``+'' configuration and approximately 404m/s$^2$ corresponding to a peak impact force of 449N in the ``$\times$' configuration. This is depicted in the bar plots, as seen for the five different trials, in Fig. \ref{fig:boxplot}A. These high impact forces were experienced by the chassis because the rigid frame does not deform, transmitting the entire impact force to the main body.
% not really distinguishing between the +' or `$\times$' configurations. 
We hypothesize that the large variance in impact force readings is due to the impact times being so miniscule, that, at the current sampling rate of the sensor (maximum of 1kHz), it would sometimes miss the peak acceleration reading. On the other hand, the soft-bodied frame of SoBAR generated deformation upon collision, which led to much longer impact time (10$\times$ in both configurations) compared to the rigid frame. 
The soft-bodied frame is capable of extending the contact impact time through its body deformation, which leads to lower impact force as shown in Figs. \ref{fig:boxplot}A and B. 
% With the peak force applied over a longer interval, the peak accelerations are noticeably smaller

Between the ``+'' and ``$\times$'' configurations of the soft-bodied frame, we notice that the ``+'' configuration experiences lower impact forces overall. Although the ``$\times$'' configuration highlights longer impact times, as seen in Table~\ref{tab:collision}, its arms mitigate the impact by splitting outwards which can be less ideal for collision mitigation. For example, Fig. \ref{fig:boxplot}A highlights the experiment with the SoBAR frame that experienced the largest recorded impact forces at 138kPa and 50cm, also captured in Fig. \ref{fig:boxplot}C(ii). We notice that at 120ms, the arms have completely split outwards, causing the chassis to impact the ground, thus leading to high impact forces. This behavior is even worse with the 69kPa frame when dropped at 50cm, saturating the on-board accelerometer.
% and therefore we did not continue the tests for 5 trials as to not damage the system. 
The slow motion capture of this test is highlighted in Fig.~S4. This behavior is analogous to Euler springs where beyond the maximum compression distance, the entire impact force is transmitted to the chassis. However, by increasing the internal pressure of 207kPa, even at the 50cm drop, the ``$\times$'' configuration is able to successfully mitigate the impact without the main chassis contacting the ground.

We further compare the collision resilience ability of SoBAR with the many available collision resilient aerial robots in Appendix~J and Supplementary Table 5. 

Since, indoor aerial robots are prone to collisions with impact velocities upto 2m/s, we chose to utilize the ``+'' configuration at 207kPa, to maximize the collision mitigation ability and minimum thrust loss (discussed in Sec.~\ref{sec:beam_bend}) for the collision and flight demonstrations with SoBAR, in Secs.~\ref{sec:collision_demo} and~\ref{sec:real_time_demo}.

%%%%%%%%%%%%%%%%%%%%%%%%%%%%%%%%%%%%%%%%%%%%%%%%%%%%%%%%%%%%
\begin{table}[t!]
	\centering
	\caption{Impact times for SoBAR and the rigid DJI F450 frame in various configurations}
	\begin{tabular}{*6c}
		\toprule
		\multicolumn{1}{c}{Frame} & \multicolumn{1}{c}{Pressure} &  \multicolumn{2}{c}{`x' configuration} &  \multicolumn{2}{c}{`+' configuration} \\
		\multicolumn{2}{c}{} &  \multicolumn{2}{c}{impact time (ms)} &  \multicolumn{2}{c}{impact time (ms)} \\
		% \midrule
		{}&{} & 25cm & 50cm & 25cm & 50cm \\
		\midrule
		Rigid & - & 22.1 & - & 8 & 8 \\
		Soft & 69kPa & 96.4 & 124.5 & 80.3 & 78.3 \\
		Soft & 138kPa & 164.7 & 84.3 & 68.3 & 122.5 \\
		Soft & 207kPa & 146.6 & 108.4 & 72.3 & 78.3\\
		\bottomrule
	\end{tabular}
	\label{tab:collision}
	\vspace{-2em}
	
\end{table}
%%%%%%%%%%%%%%%%%%%%%%%%%%%%%%%%%%%%%%%%%%%%%%%%%%%%%%%%%%%%
%%%%%%%%%%%%%%%%%%%%%%%%%%%%%%%%%%%%%%%%%%%%%%%%%%%%%%%%%%%%

%%%%%%%%%%%%%%%%%%%%%%%%%%%%%%%%%%%%%%%%%%%
\subsubsection{Collision Demonstration}
\label{sec:collision_demo}
To verify SoBAR's collision performance, we carried out a series of experiments where the drone took off and approached the target setpoint without the knowledge of the wall. Upon collision, it recovered and the collision trajectory was recorded in slow motion, as seen in Fig.~\ref{fig:boxplot}F. 
It is seen that due to the deformation of the soft body, low rebound velocities in the range of 1.5m/s corresponding to high collision velocities of up to 2m/s are achieved.
% , such that the effective coefficient of restitution is around 0.75. 
These low rebound velocities help in post-collision recovery control without the need for complex collision characterization \cite{patnaik2021collision}. The head-on collision performance is highlighted in Supplementary Video 4.
% as described in Appendix \ref{appendix:collision_planning} . 
For the conventional rigid chassis, the rebound velocities are significantly higher, \textcolor{black}{as observed through our experimental tests that were also} captured with the high speed camera and shown in Supplementary Video 2 and Appendix~\ref{appendix:all_collision_drop_tests}, leading to complex collision characterization and recovery control \cite{patnaik2021collision}.

%%%%%%%%%%%%%%%%%%%%%%%%%%%%%%%%%%%%%%%%%%%

%%%%%%%%%%%%%%%%%%%%%%%%%%%%%%%%%%%%%%%%%%%%%%%%%%%%%%%%%%%%%
\subsection{Perching with SoBAR}
\label{sec:perch_eval}

%%%%%%%%%%%%%%%%%%%%%%%%%%%%%%%%%%%%%%%%%%%
\begin{figure*}[t!]
	\centering
	\includegraphics[width =0.97\textwidth]{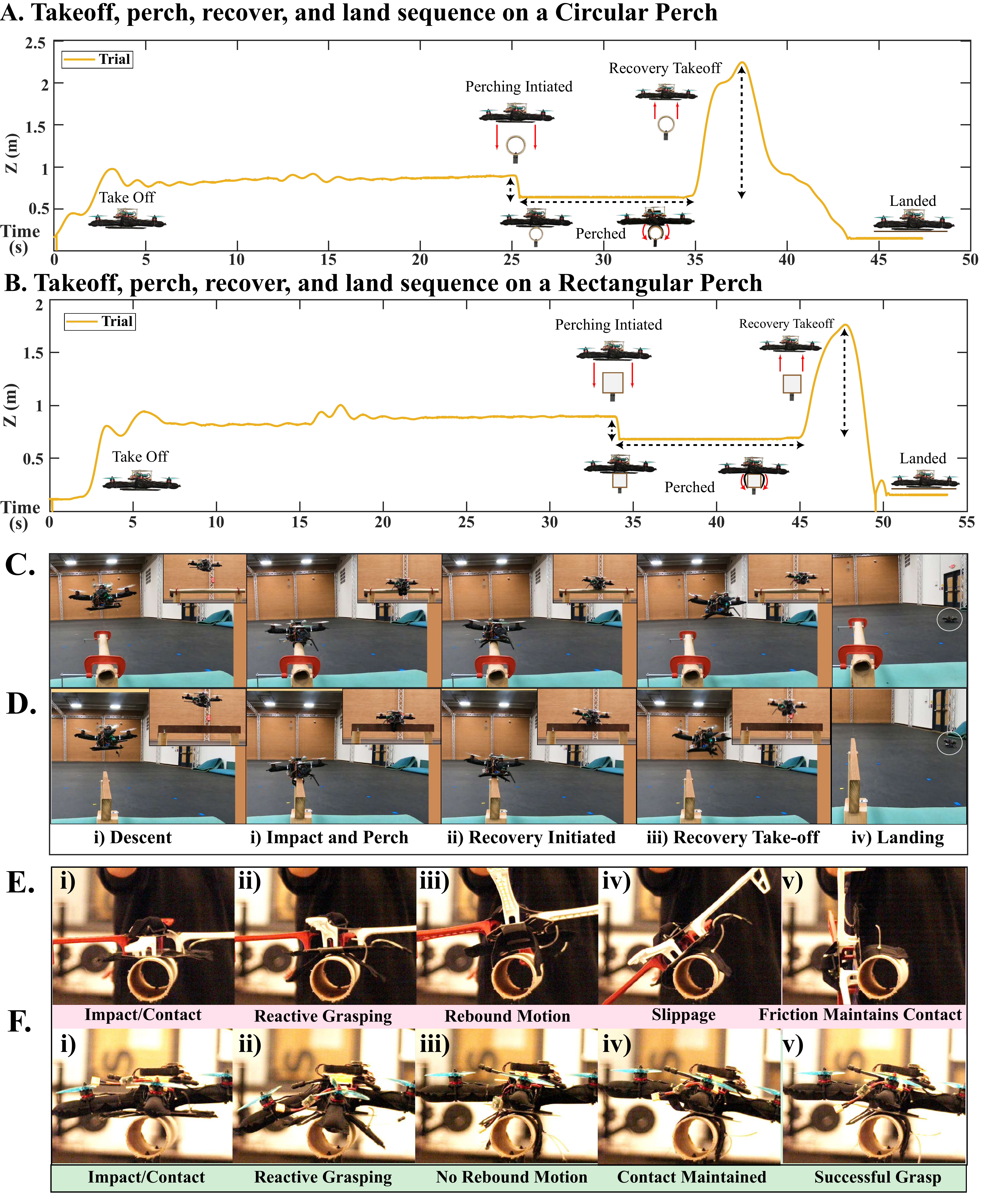}
	\caption{Takeoff, perch, recover, and land sequence of SoBAR for one of the three successful trials. (A) Successful perching on a circular object can be inferred from SoBAR's `z' position trajectory. (B) Successful perching on a rectangular object can be inferred from SoBAR's `z' position trajectory. (C) Real-time perching sequence of soft-bodied aerial robot on circular perch, from front and side view. (D) Real-time perching sequence of soft-bodied aerial robot on rectangular perch, from front and side view. (E) Rigid frame with perching mechanism demonstration. Notice a large rebound motion at (ii) causing slippage in (iv). (F) Soft aerial robot with perching mechanism demonstration}
	% \subfloat[Wrench Hull Obtained]{\includegraphics[width = 0.4\textwidth]{}}
	\label{fig:perch_cir_rect}
	\vspace{-1em}
	
\end{figure*}
%%%%%%%%%%%%%%%%%%%%%%%%%%%%%%%%%%%%%%%%%%%%%%%%%%%%%%%%%%%%%

\subsubsection{Real-time Experiments}
\label{sec:real_time_demo}
The $z$ position trajectory of SoBAR for real-time perching is shown in Figs. \ref{fig:perch_cir_rect}A and B. This coordinate was chosen to mark the various phases of the maneuver, as shown in the Figs. \ref{fig:perch_cir_rect}A and B. We demonstrate SoBAR's perching ability on a cylindrical and rectangular cross-sectioned objects, also highlighted in Supplementary Video 5. Initially, SoBAR approaches and hovers above the target location, which is obtained by placing an infrared marker on the perching object. This hover phase is marked where the height is oscillating due to small errors in the tracking control. Once the position errors fall within the preset bound, the free-fall descent is initiated for perching, this is shown by the steep slope of the $z$ position trajectory. After the successful perch, SoBAR rests for a user defined wait time, and then performs a manual recovery and landing sequence. We conducted three autonomous perching experiments for two different objects and the perching results for one run on each object are shown in Figs. \ref{fig:perch_cir_rect}A and B, respectively. As previously mentioned, since the object radius is within the maximum grasp radius, we have significantly high success rates (four out of five times). Figures \ref{fig:perch_cir_rect}C and D, highlight snapshots of the experimental setup and the real-time perching, recovery, and landing maneuvers.

%%%%%%%%%%%%%%%%%%%%%%%%%%%%%%%%%%%%%%%%%%%%%%%%%%%%%

\subsubsection{Perching with Rigid vs. Soft-Bodied Frame}
We compared the perching ability between SoBAR and rigid DJI frames, mounted with the HFB grasper, dropped from a height of 25cm onto a circular perch, as seen in Figs.~\ref{fig:perch_cir_rect}E and F, and Supplementary Video 5. For the rigid frame, we notice that the impact times are significantly smaller (2-4ms) leading to higher impact forces. These forces are too large to not only activate the grasper, but also cause a rebound motion, as shown in Fig.~\ref{fig:perch_cir_rect}E(ii) and (iii), which causes the system to bounce off the perch, before the grasper can successfully engage, as shown in Fig.~\ref{fig:perch_cir_rect}F(iii). As the grasper is already partially activated, the friction on the fingers helps the rigid frame from completely falling off. However, because of the uneven application of wrench forces, there is noticeable slippage, seen in Fig.~\ref{fig:perch_cir_rect}E(iv), until enough contact is reached in Fig.~\ref{fig:perch_cir_rect}E(v). This leads to unpredictable final configurations, as shown in Fig.~\ref{fig:perch_cir_rect}E(v) (roll angle almost 90$^o$), which can be difficult to recover from. The success rate for this case scenario was one out of five. 

We notice that upon impact SoBAR's frame undergoes deformation thus increasing the impact duration, as shown in Figs.~\ref{fig:perch_cir_rect}F(iii) to (v). This allows contact with the perch for a longer time, leading to successful engagement of the HFB grasper. SoBAR was able to successfully perch four out of five times. This highlights how the deformable body improves the perching capabilities in comparison to the rigid frame, and aids in the application of the wrench forces, without explicit grasp hull computations and results in a robust grasp during the perching task.

We perform further characterization of the grasping performance by running situational experiments with a rigid frame, rigid frame with additional mechanical damping and the SoBAR frame at various internal pressures as shown in Appendix G1 and Supplementary Video 6. The success rates are documented in SI Table 2. In all cases, it can be verified that the SoBAR frame can successfully engage the gripper and provide high success rates.

\textcolor{black}{We also extend the drop tests of SoBAR to other real-world objects, such as a hard-hat helmet (22cm diameter), edge of a ladder (4x2cm), a rock (6.5-8cm width, 23cm height), a tree branch (7.3cm diameter), a joint of a UR-5 robot arm (10cm diameter), and a sanitizer stand (13x18x10cm), highlighting the capabilities of the system to perch onto objects with different shapes and sizes, as seen in Appendix~G2 and Supplementary Video 7.}

We characterize the stability of the perch with SoBAR and the HFB grasper against various other state-of-the-art perching aerial robots in Appendix I and Supplementary Table 3. We note that, SoBAR is capable of perching on irregular objects, cylindrical objects and also on planar surfaces when compared to other proposed avian-inspired graspers.

Finally, we demonstrate in Supplementary Video 8, how SoBAR can successfully perch while undergoing collisions from a nearby wall. The rigid frame with or without the damper failed in all simultaneous wall-collision tests during perching. 
	% This is attributed to two main reasons, the approach angle is such that the gripper engages, however the rebound velocity from the wall collision is high enough to let the vehicle slip sideways before the perch is successful. Secondly, the approach angle is such that the gripper is not aligned to the target due to the non-conformable nature of the rigid arm. 
	The SoBAR frame, at higher stiffness (at internal pressure 103kPa and 138kPa), was not successful on every trial. But, the SoBAR frame at 70kPa allowed deformation of its arm upon impact absorbing the collision forces and simultaneous perching successfully. Further details are documented in Appendix M.

%%%%%%%%%%%%%%%%%%%%%%%%%%%%%%%%%%%%%%%%%%%%%%%%%%%%%%%%%%%%%
%%%%%%%%%%%%%%%%%%%%%%%%%%%%%%%%%%%%%%%%%%%%%%%%%%%%%%%%%%%%%%%%%%%%%%%%%%%%%%%%%%%%%%%%%%%%%%%%%%%%%%%%%%%%%%%%

\section{Conclusion}
\label{sec:conclude}
\subsection{Summary}
This paper presented the design, development, and evaluation of an untethered, lightweight, robust, and compliant fabric soft-bodied aerial robot (SoBAR) composed of a soft-bodied frame and hybrid fabric-based bistable (HFB) grasper. \textcolor{black}{The vision of this work was geared towards addressing two previously disconnected capabilities in aerial robots, namely, impact mitigation and dynamic perching, through the co-development of the body and grasper of the aerial robot. 
}

The SoBAR frame and the HFB grasper operate synergistically to perform high-speed, high-impact, and dynamic collision-based perching. We observed that the SoBAR had a perching success rate of four out of five times on objects within the grasp radius of the HFB grasper during our experiments. Furthermore, we also demonstrate how SoBAR can successfully perch while undergoing a wall-collision, by exploiting its deformable arms and variable stiffness.

\subsection{Limitations and Future Directions}
\textcolor{black}{As with every novel design, the current SoBAR comes with some observed limitations, which are open challenges to future iterations for improvement. To start with, we experienced a beam deflection of approximately 5.80$^\circ$ at 207 kPa which led to slight thrust loss (maximum loss coefficient of 0.95 at 69kPa) and force residuals in the body $x-y$ plane. Thrust loss was addressed in the controller design by modeling the thrust coefficients as a function of the deflection angle and by selecting the motor-propeller pair that produces higher thrust-to-weight ratio for compensation. In the future, we aim to maximize the energy efficiency of the soft body by optimizing its design for improved stiffness at various internal pressures. This can be achieved through varying the design of its cross section\cite{sridar2017development}, including internal soft truss structures, and/or developing a rigid-soft hybrid frame. The currently utilized Euler-beam theory for calculating the thrust coefficients is less accurate in the presence of shear stresses at lower internal pressures. To address this, we will explore detailed modeling of the pneumatic beam and the corresponding thrust vectoring via machine learning methods for developing model-based low-level flight controllers.} 

\textcolor{black}{Similarly, for the HFB grasper, although we observed that it was adaptable to many irregular shapes, we feel there still is room for design improvements. Currently, the HFB graspers are mounted against a flat plate, which slightly limits their bending range. This therefore reduces the maximum surface contact between the grasper and the perching object, as the region of the grasper mounted to the flat plate cannot bend or curl. We anticipate that by redesigning the grasper mount to fully utilize its continuum nature, its entire grasping surface area would be utilized more effectively. Furthermore, through exploration of different bistable materials with varying thicknesses has the potential in improving the  torsional stability of the grasper. The above design improvements combined with pre-strained adaptations of parts of the grasper can further increase its grasping force. Finally, exploration in the addition of adaptive surface structures, such as microspines or microhooks, could increase its friction coefficient and thus grasping performance.}
 
During the collision resilience validation in the body $x-y$ direction via head-on collisions,
% we demonstrated a successful recovery of the SoBAR from a speed up to 2m/s. However, 
we noticed that after a long collision test session, the motor mounts on the arms of SoBAR, seen in Fig.~\ref{fig:fabrication}A(ii) had to be readjusted due to minor slippage. This can be addressed by adding anti-slip fabric at the interface between the motor mount and the fabric frame. In addition, we could also sew the motor mounts directly on the fabric frame, with minor sacrifice of stowability.

For immediate future work, we seek to demonstrate real-life highly dynamic autonomous perching with SoBAR by visually detecting suitable perches.
\textcolor{black}{
	Aggressive trajectory planning and robust control strategies will also be explored that can enable approaching objects at different angles and heights in an obstacle laden environment. 
} 

The discussed future efforts have the potential to enable battery recharging techniques for prolonged outdoor monitoring, as well as search and data collection missions. 

Finally, our work lays a stepping stone to design soft and compliant aerial robots where the chassis is utilized for achieving intrinsic safety and collision resilience. We hope that the insights from our work shall inspire novel bio-inspired designs for soft reconfigurable and conformable aerial-robots to obtain various functionalities such as  whole-body perching and grasping.

%%%%%%%%%%%%%%%%%%%%%%%%%%%%%%%%%%%%%%%%%%%%%%%%%%%%%%%%%%%%%%%%%%%%%%%%%%%%%%%%
%%%%%%%%%%%%%%%%%%%%%%%%%%%%%%%%%%%%%%%%%%%%%%%%%%%%%%%%
%TC:ignore

\section*{ACKNOWLEDGEMENT}
We would like to thank Saivimal Sridar, Weijia Tao, YiZhuang Garrad,  Aravind Adith and Sunny Amatya for their help with fabrication, setting up video recordings, setting up the UR5 tests, and helpful suggestions during the ideation and paper editing process.

%TC:endignore

%%%%%%%%%%%%%%%%%%%%%%%%%%%%%%%%%%%%%%%%%%%%%%%%%%%%%%%%%%%%%%%%%%%%%%%%%%%%%%%%
%TC:ignore
\bibliographystyle{ieeetr}
\bibliography{references.bib}
%TC:endignore

%%%%%%%%%%%%%%%%%%%%%%%%%%%%%%%%%%%%%%%%%%%%%%%%%%%%%%%%%%%%%%%%%%%%%%%%%%%%%%%%%%%%%%%%%%%%%%%%%%%%%%%%%%%%%%%%%%%%%%%%%%%%%%%%%%%%%%%%%%%%%%%%%%%%%%%%%%%%%%%%%%%%%%%%%%%%%%%%%%%%%%%%%%%%%%%%%%%%%%%%%%%%%%%%%%%%%%%%%%%%%%%%%%%%%%%%%%%%%%%%%%%%%%%%%%%%%%%%%%%%%%%%%%%%%%%%%%%%%%%%%%%%%%%%%%%%%%%%%%%%%%%%%%%%%%%%%%%%%%%%

\newpage

%TC:ignore
\appendix
	
	\subsection{Fabrication of SoBAR Frame and HFB Grasper}
	\label{appendix:fabrication}
	A unibody structure was employed to fabricate SoBAR's frame, as seen in  Fig. 2A. The nylon fabric, parchment paper, and TPU material (DT-2001, American Polyfilm, Branford, CT) were first cut into the desired morphology using the laser-cutter \textcolor{black}{(Glowforge Inc., Glowforge, Seattle, WA)}, shown in  Fig. 2A(i). The TPU bladder was made by aligning two TPU sheet cut-outs, sandwiching the parchment paper cut-out in the middle, and heat-sealed utilizing the (FLHP 3802, FancierStudio, Hayward, CA), at 275$^{\circ}$F for 45s. The pneumatic fitting (5463K361, McMaster-Carr, Elmhurst, IL) was also added in the TPU bladder. The two sheets of nylon fabrics were sewn along the edges, utilizing a super-imposed seam, and the complete TPU bladder was inserted in the middle of the prepared nylon fabric shell, to complete SoBAR's frame, as seen in  Fig. 2A(ii).
	
	In order to fabricate these HFB actuators, we wanted to utilize a lightweight bistable material that would maintain a straight beam state but also is capable of switching to a curled state upon contact with the perch, inspired by the snap-bracelets seen in kid's toys. To utilize a \textcolor{black}{low-cost} off-the-shelf solution, we chose the bistable metallic tape-spring, that would allow us to scale the length of the actuator as well as its thickness, by \textcolor{black}{stacking} multiple tape-springs. We first cut the measuring tape \textcolor{black}{(STANLEY STA030696N, Stanley Inc., Arlington County, VA)} to the desired size, seen in  Fig. 2B(i). We also chamfered the edges for safety. The bistable metal has two sides, with one being concave and the other convex. To pre-form the spring steel, we rolled and bent it tightly along the convex side, around a cylindrical object. The tightly curled spring steel was wrapped to maintain shape, for
	\textcolor{black}{a minimum of} 30min, as shown in  Fig. 2B(ii). The spring steel was then able to switch between two states: (i) straight beam (ii) curled state, in  Fig. 2B(iii). 
	
	The TPU material, parchment paper, nylon fabric, and 210D TPU-coated nylon fabric (DIY Packraft Ltd., Smithers, BC), were cut utilizing a laser cutter, as in  Fig. 2B(iv). A TPU actuator was manufactured in order to perform recoil after perching. Three pre-formed tape spring steels were aligned and sandwiched between the TPU-coated nylon sheets, and heat-sealed with the heat press, to make the spring steel set, shown in  Fig. 2B(iv). A pouch was then made utilizing nylon fabric, by sewing the edges and the TPU actuator and spring steel set were inserted in the pouch. Finally, the bottom surfaces of each grasping actuator were equipped with high-friction grip material (3M TB614, 3M Company, Maplewood, MN), completing the fabrication of the HFB actuator. Each completed actuator weighs only 38g. The multi-fingered perching mechanism can be designed in different orientations, as in  Fig. 2A(iv). In this work, we only tested the two-fingered and three-fingered configurations for the soft-bodied aerial robot, as depicted in  Fig. 2A(iii).  
	
	%%%%%%%%%%%%%%%%%%%%%%%%%%%%%%%%%%%%%%%%%%%%%%%%%%%%%%%%%%%%%%%%%% {c{3.75cm} c{1.25cm} c{2.2cm}}
	% \begin{table}
	% \centering
	% \caption{Mass Budget of SoBAR}
	% \label{tab:mass_budget}
	%     \begin{tabular}{p{3.75cm}c{2cm}c{4cm}}
	%     \textbf{Unit} & \textbf{Weight (g)} & \textbf{Total Weight Percentage (\%)} \\ \hline
	%   Soft-bodied frame & 10 & 0.88 \\
	%   4 $\times$ (motor and propeller pair) & 126 & 11.09 \\ 
	%   Micro diaphragm pump & 100 & 8.79 \\
	%   4S LiPo battery & 250 & 21.99 \\
	%   Chassis & 81 & 7.12 \\
	%   4 $\times$ (motor mounts) & 128 & 11.26 \\
	%   Flight controller & 50 & 4.4 \\
	%   Intel UP board & 100 & 8.79 \\
	%   2 $\times$ (grasper fingers) & 76 & 6.68 \\
	%   Grasper mount & 39 & 3.43 \\
	%   Other electronics (power supply, BEC mo.
	% *dules, voltage regulator, wifi module etc.) 
	%     other (miscellaneous) & 177 & 15.57 \\ \hline
	% \end{tabular}
	% \end{table}
	%%%%%%%%%%%%%%%%%%%%%%%%%%%%%%%%%%%%%%%%%%%%%%%%%%%%%%%%%%%%%%%%%
	\begin{table}
		\centering
		\caption{Mass Budget of SoBAR}
		\label{tab:mass_budget}
		\begin{tabular}{p{4cm}p{1.5cm}p{1.5cm}}
			\textbf{Unit} & \textbf{Weight (g)} & \textbf{Total Weight Percentage (\%)} \\ \hline
			Soft-bodied frame & 10 & 0.9 \\
			4 $\times$ (motor and propeller pair) & 126 & 11 \\ 
			Micro diaphragm pump & 100 & 8.8 \\
			4S LiPo battery & 250 & 22 \\
			Chassis & 81 & 7.1 \\
			4 $\times$ (motor mounts) & 28 & 2.4 \\
			Flight controller & 50 & 4.4 \\
			Intel UP board & 100 & 8.8 \\
			2 $\times$ (grasper fingers) & 76 & 6.7 \\
			Grasper mount & 39 & 3.4 \\
			Other electronics (power supply, upboard mounts, BEC modules, voltage regulator, wifi module and 
			other miscellaneous) & 277 & 24.5 \\ \hline
		\end{tabular}
	\end{table}
	%%%%%%%%%%%%%%%%%%%%%%%%%%%%%%%%%%%%%%%%%%%%%%%%%%%%%%%%%%%%%%%%%%

	%%%%%%%%%%%%%%%%%%%%%%%%%%%%%%%%%%%%%%%%%%%%%%%
	\subsection{Electronics, Mass Budget and Integration}\label{appendix:integration}
	SoBAR's chassis hosts the flight controller, power module, and high level controller, as seen in Fig. 3. Flight controller utilized is a PIXHAWK flight controller with the Intel UP Board as the companion computer. The high-level companion computer is used to relay the position and orientation data from the indoor motion capture system to the flight controller at 120 Hz for objects of interest within the flight arena (SoBAR and the perching objects). 
	Analog pressure sensors (ASDXAVX100PGAA5, Honeywell International Inc., Morris Plains, NJ) and a micro diaphragm pump (NMP830 HP-KPDC-B) are used to control the pressure of the body and the HFB actuator. The PIXHAWK flight controller is modified from the off-the-shelf code in order to integrate the micro-pump, its controller, and control allocation. The onboard micro-pump is connected to the soft grasper and body frame. In this work, the soft-bodied frame is inflated up to 207kPa and evaluated at intervals of 69kPa. To fully recoil the soft grasper only 83kPa is required. The flight control unit connects to the pressure sensor and micro-pump using I$^2$C and Analog-Digital (AD) interfaces respectively. A standard proportional controller is implemented to control the pressure output from the micro-pump. \textcolor{black}{By changing the desired setpoint of the pressure, this single pump has been used to inflate the body and to  disengage the HFB grasper after perching for recovery. }A 4S lithium polymer battery of 3300 mAh LiPo battery of 14.8V, 50C is used for the power supply. The motors are controlled utilizing Lumenier 30A BLHeli\_S Electronic Speed Controllers (ESCs) and the entire system has a maximum thrust-to-weight ratio of 4.58:1. The mass budget of the system is highlighted in Table~S\ref{tab:mass_budget}. Noticeably, the soft robotic components and their mounting brackets make up only 19.7\% of the entire system. Overall, SoBAR has a size of 319$\times$319mm and weighs 1.14kg. 
	
	%%%%%%%%%%%%%%%%%%%%%%%%%%%%%%%%%%%%%%%%%%%%%%%
	\subsection{Experimental Setup}
	\label{appendix:exp_setup}
	To characterize the HFB actuators and SoBAR’s deflection, a universal tensile testing machine (UTM) (Instron 5944, Instron Corp., High Wycombe, United Kingdom) was utilized. To monitor the collision and drop tests, and the HFB actuator activation time, a 500fps high-speed camera (Edgetronics SC1, CA, USA) was utilized. 
	% % \kay{
	% \subsubsection{Grasper Slip Detection}~\\
	% Initially when the UTM pulls on the cylindrical object, the grasper exerts forces to retain the grasp. However, as the UTM continuously increases the pulling force, at one point the grasping force starts decreasing indicating a slip condition for the grasp. The maximum value observed on the UTM is noted to characterize the maximum grasping force of the grasper on that particular object, as shown in Fig.~\ref{fig:grasping_force}.} 
	To set up the drop tests, the frames were mounted on the UR5 robot manipulator, with a controlled Hand-E grasper (Universal Robotics, Odense, Denmark). To monitor the peak impact accelerations of the drop tests, a high-G accelerometer SparkFun H3LIS331DL (Sparkfun, Boulder, Colorado), with a maximum reading of 400G, was utilized. The sensor was capable of measuring acceleration at 1kHz sampling rate. Finally, the indoor perching experiments were performed utilizing a Vicon motion capture system (OptiTrack, NaturalPoint, Inc., Corvallis, OR) to obtain the position and orientation information of SoBAR and the perching location. SoBAR’s soft-bodied frame’s stiffness was varied by modifying the internal pressure increments of 69kPa, from 69kPa up to 207kPa, throughout the experiments.
	
	% The peak accelerations were measured with an onboard high-G accelerometer. SoBAR’s soft-bodied frame’s stiffness was varied by modifying the internal pressure increments of 68.9kPa, from 68.9kPa up to 204.8kPa. The dimensions and onboard weight (at 1.1kg to represent the actual weight of SoBAR) was maintained between both frames. 
	
	%%%%%%%%%%%%%%%%%%%%%%%%%%%%%%%%%%%%%%%%%%%%%%%
	
	\subsection{Perching Task Planning}\label{appendix:planning}
	\label{appendix:perching_planning}
	Due to the complexities in autonomous recovery control and due to the scope of this paper, we present autonomous perching with a manual recovery control, leaving autonomous recovery for future work. 
	
	The entire perching maneuver consists of multiple control strategies which are described in this section. The first trajectory involves a maneuver where SoBAR's flies to the perching location to hover till the error in position is near zero and descends over the perching target with a specified downward velocity. This velocity is computed from a drop test by iterating over the height $h$ which engages the grasper to achieve a successful perch. This also corresponds to the activation force (as computed in Sec. 3.1) for a impact time of about 0.1s. Neglecting air resistance for low velocities, the impact velocity is calculated using 
	\begin{equation}\label{eqn:impactvel}
		\centering
		\begin{aligned}
			v_t = \sqrt{2gh}
		\end{aligned}
	\end{equation}
	where $v_t$ is the impact velocity and g is the constant for acceleration and $h$ is the height from which the platform is dropped.
	We see that, for the current system weight, the impact force generated by a free fall from a minimum height of 30cm is effective to engage the a three finger HFB grasper successfully and the corresponding impact velocity is approximately 2.4m/s. These values are around 20cm corresponding to 1.98m/s for a two finger HFB grasper. The reference trajectory for the downward descent are therefore chosen as the $x-y$ coordinates of the perching target and the $z$-direction velocity for the cascaded P-PID low level position controller. The perching maneuver strategy is shown in the orange colored sub-block of the block diagram in Fig. 4.

	\subsection{Single HFB Actuator Evaluation}
	\label{sec:grasp_eval_appen}
	
	\subsubsection{Single Actuator Activation and Recovery Time}
	% In this test, we evaluated the time taken for each finger to go from a straight beam state to a curled state (activation) and pneumatically recover the actuator from the curled state back to its initial state, as shown in Fig. 1B and Supplementary Figure S\ref{fig:final_collage}A. To do so, we tracked the time sequence using a high-speed camera at 500fps. We set the pressure to recoil back to a straight beam state at 83kPa, which is the minimum pressure required to recoil from prior trial-and-error testing. From the high-speed footage, we measured the activation stage to take only 4ms and the pneumatically actuated recovery stage to take 3s. This is highlighted in the Supplementary Video 2 as well.
	In this test, we evaluated the time taken for each finger to go from a straight beam state to a curled state (activation) and pneumatically recover the actuator from the curled state back to its initial state, as shown in Fig. 1B and also Figure 1F. To do so, we tracked the time sequence using a high-speed camera at 500fps. We set the pressure to recoil back to a straight beam state at 83kPa, which is the minimum pressure required to recoil from prior trial-and-error testing. From the high-speed footage, we measured the activation stage to take only 4ms and the pneumatically actuated recovery stage to take 3s. This is highlighted in the Supplementary Video 2 as well.
	%%%%%%%%%%%%%%%%%%%%%%%%%%%%%%%%%%%%%%%%%%%%%%%%%%%%%%%%%

	\subsubsection{Single Actuator Tip Force}
	
	A single HFB actuator was utilized to evaluate the actuator tip force, using the UTM. The finger’s tip is placed in contact with the UTM's load cell. The finger is then activated at the proximal end, and the tip force is measured. The actuator tip force with a single spring steel is 0.16N and that with three embedded spring steels is 0.55N.
	
	%%%%%%%%%%%%%%%%%%%%%%%%%%%%%%%%%%%%%%

	\subsubsection{Single Actuator Activation Force}
	
	To evaluate the force required to activate a single actuator (the activation force required for the actuator to switch from a straight beam state to a completely curled state), we aligned a single actuator underneath the load cell of the UTM, with the convex side up.
	% , as seen in Fig.~\ref{}. 
	The UTM was designed to push downwards at a rate of 8mm/s until the activation force is recorded. The activation force of 7N, 24N, and 54N was recorded for an actuator with a single, triple, and quintuple embedded spring steels, respectively. 
	
	From the activation force, we are able to calculate the approximate desired impact velocity. The impact time is approximately 0.1s as visualized by the high-speed camera, leading to impact velocities of 0.7m/s, 2.4m/s, and 5.4m/s in the body $z$ direction of SoBAR, for the single, triple, and quintuple embedded spring steels graspers, respectively. For real-world experiments, we decided to utilize the grasper with three embedded spring steels by optimizing the three parameters of required impact velocity, achieved grasp force and total weight addition by the grasper.
	%However, as long as sufficient forces can be applied, the grasper can be used to exploit its ability to exert significantly high wrench forces for other applications.
	
	%%%%%%%%%%%%%%%%%%%%%%%%%%%%%%%%%%%%%%%%%%%%%%
	%%%%%%%%%%%%%%%%%%%%%%%%%%%%%%%%%%%%%%%%%
	\subsection{Static Wrench Analysis of the Grasper for Non-Conformable Objects }
	\label{sec:wrench_analysis}
	
	%%%%%%%%%%%%%%%%%%%%%%%%%%%%%%%%%%%%%%%%%%%%%%%%%%%%%%%%%%%%%%%%%%%%%%%%%%%%%%%%%%

	%%%%%%%%%%%%%%%%%%%%%%%%%%%%%%%%%%%%%%%%%%%%%%%%%%%%%%%%%%%%%%%%%%%%%%%%%%%%%%%%%%%%%%%%%%%%%%%%%%%%%%
	
	We perform a static grasp wrench analysis \cite{prattichizzo2016grasping}
	to obtain an insight into whether SoBAR with the HFB grasper will successfully perch on objects, with shapes and sizes, that do not conform to the grasper workspace. For the wrench analysis, we assume a 2D object and the wrench space in $\mathbb{R}^3$, neglecting any movement in longitudinal direction of the perching object.
	% This static grasping analysis of the grasper for objects that do not conform to its workspace, helps us further understand robustness of the grasp in real perching scenarios.
	
\begin{figure}[t]
		\centering
		\includegraphics[width=0.4\textwidth]{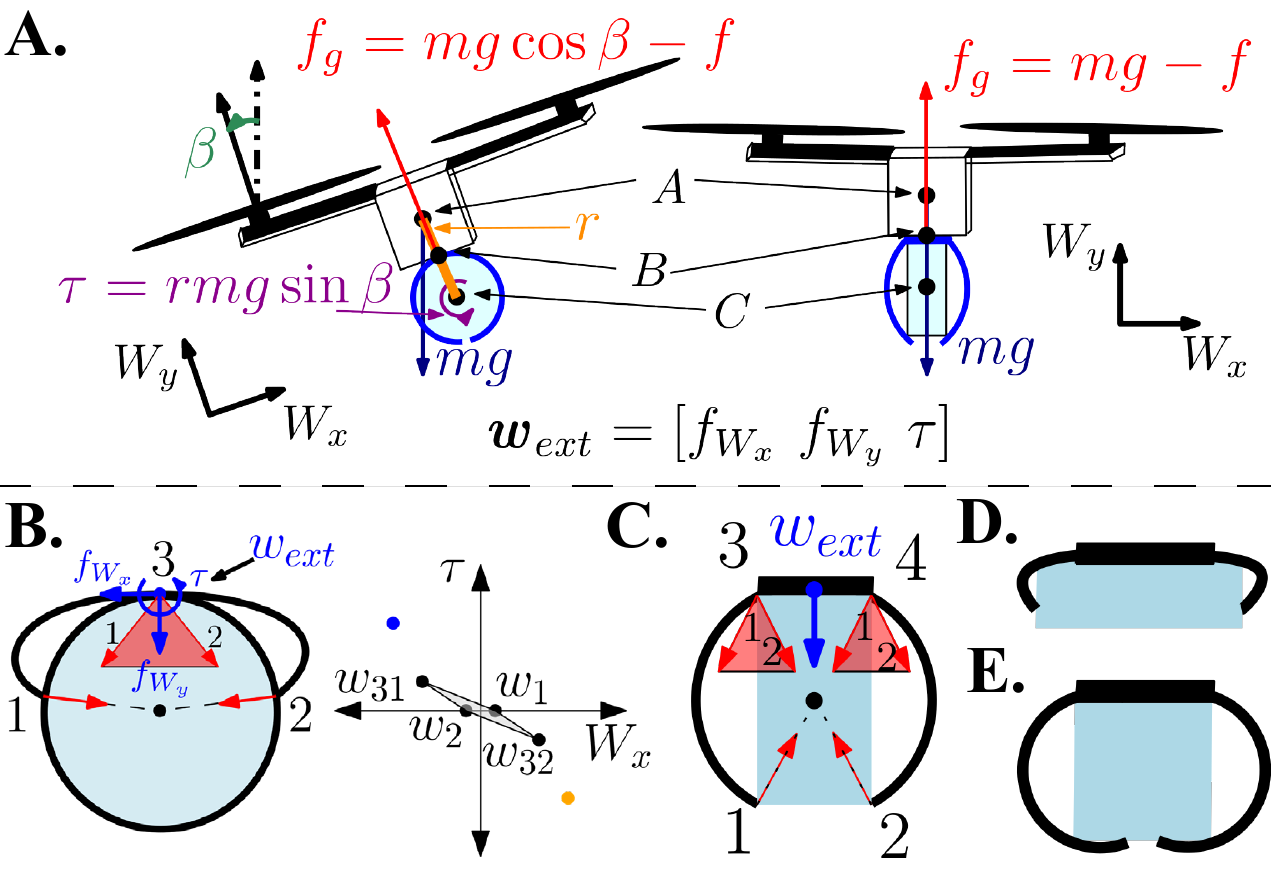}
		% \setlength{\abovecaptionskip}{3pt}
		% \setlenlowcaptionskip}{-10pt}th
		\caption{Modeling the various features of SoBAR for perching and flying conventionally. (A) External wrench for circular and square objects. (B) Wrench hull for circular object larger than the grasp radius. (C) Grasp consisting of one friction force and two frictionless point forces (D) Perching on a side larger than than grasp radius (E) Perching on an object within the grasp radius.}
		\label{fig:gwh}
		\vspace{-0.3in}
		% \vspace{-1.5em}
	\end{figure}

	Supplementary Figure~S\ref{fig:gwh}A illustrates the external wrench on the grasper in different scenarios (objects with circular and rectangular cross-sections) and the reference wrench frame, $W$, which is located at the center of the grasper and oriented with the $b_3$ axis. The gravitational force ($mg$) acting on SoBAR's’s center of gravity (Point A) and the reaction force ($f_g$) acting at the point of contact (Point B) is transferred to Point C, the center of the object. Note that the reaction force compensates for any residual thrust from the motors, denoted in the figure by $f$.  
	% The grasp is modelled by three contact forces in both scenarios shown by points 1,2 and 3 in both \ref{fig:gwh}B and \ref{fig:gwh}C. 
	For the circular cross-section, there is an acting torque at Point C due to the perch orientation but for the rectangular cross-section, the external wrench will only consist of the contact forces and gravitational forces, excluding any torques, due to the flat surface on which SoBAR perches. In order to ensure a successful perching, the maximum force by the grasper should be able to generate an equal wrench in the opposite direction. For the analysis that follows, the friction coefficient, $\mu$, between the high-friction grip material
	% \cite{3M} 
	used on the grasper and the cardboard material of the perching object is taken as 0.7 as calculated from experiments.  
	% according to the data sheet provided~\cite{3M}.
	
	% reference link and data sheet for high-friction grip material 
	% https://www.amazon.com/3M-TB531BLK-grasping-Material-Length/dp/B00OPSTXKQ/ref=pd_sbs_3/130-1112615-0639251?pd_rd_w=fOr1M&pf_rd_p=0a3ad226-8a77-4898-9a99-63ffeb1aef90&pf_rd_r=ZFGAXY5WHM644ST8YXE4&pd_rd_r=771f928d-9a3d-44f2-919a-56ded5fcd818&pd_rd_wg=Juqdq&pd_rd_i=B00OPSTXKQ&th=1
	% https://multimedia.3m.com/mws/media/610268O/3m-grasping-material-gm110-and-gm400.pdf
	
	We first consider a scenario where SoBAR's perches on a circular object with a diameter (115mm) greater than the workspace of the two-finger grasper (70mm). 
	% Considering the weight of MAV as 1.14kg, as seen in Table~\ref{tab:mass_budget}, a perching angle of approximately 30$^\circ$, diameter of the cylinder as 115mm, and the distance between Point A and Point B as 50mm, then the external wrench $w_{ext}$ consists of a force of 11.17N and an external torque of 0.6Nm. 
	With the values $m = 1.14\text{kg}, \beta = 30^o, g = 9.81\text{m/s}^2,$ and 
	\[ r = r_{AB} + r_{BC} = \Bigg(50 + \frac{115}{2}\Bigg)\text{mm} = 107.5\text{mm,} \]
	we can calculate the components of the external wrench, $w_{ext}$, as:
	\[f_{W_x} = -mg\sin{\beta} = -5.59\text{N},\] 
	\[ 0 \leq f_{W_y} \leq (mg \cos{\beta} - f)\] and \[\tau = 0.6\text{Nm,}\] where $f$ is residual thrust during perching and is significantly less than $mg$. To compute the grasp wrench hull, we can assume that there are three forces as shown in the free body diagram of Supplementary Figure S\ref{fig:gwh}B. The forces are approximated as follows - a friction cone at point ``3'' and two frictionless point forces at locations ``1'' and ``2'' respectively which are directed towards the center of the object. To account for the curled tip force that is not exactly directed towards the center, we approximate a loss cone of $\pm5^o$ within which the tip force lies. Depending on the grasp wrench hull, we will be able to tell if the external wrench (shown by the blue dot in Supplementary Figure S\ref{fig:gwh}B) can be compensated by the grasper in the configuration for a large object as shown in Supplementary Figure~S\ref{fig:gwh}B. With the above-mentioned parameters, we obtain:
	\[f_{g_{max}} = mg \cos{\beta} = 9.68\]
	Using the tip force values as calculated from experiments and neglecting the small torque generated by these forces:
	\[ f_1 = -f_2 = 0.55\text{N,}\] and \[f_{31} = -f_{32} = \mu f_{g_{max}} = 6.78\text{N}\] with a friction cone of angle \[ \alpha = \tanh{\mu} = 35^o\]
	We can now calculate the wrench generated by the grasper as 
	% the values of the $W_y$ component of force and torque is calculated as $w_{ext}|{W_y} = [-11.17~0.6]^T, 
	\[w_1 = [0.54 ~-0.1 ~0]^T, ~w_2 = [-0.54 ~-0.1 ~0]^T,\] \[w_{31} = [-f_{31} \sin{\alpha} ~~ f_{31} \cos{\alpha} ~~ r_{BC}f_{31} \sin{\alpha} ]^T\] \[ = [-3.88 ~-5.55 ~0.22]^T\] and similarly \[w_{32} = [3.88 ~-5.55 ~-0.22]^T\]
	We can infer that no linear combination of the grasper forces can cancel out the external wrench. As shown in the  Supplementary Figure~S\ref{fig:gwh}B, the required wrench (the yellow dot) falls outside the wrench hull in this case, and hence leads to an unstable perch. This is accounted by the fact that the grasper cannot generate required force and torque in this configuration to cancel out the external wrench's component in $W_x$ direction and $\tau$. 
	
	% t_ext = m*g*sin 30 * (50+(115/2))/1000
	
	We proceed in a similar way to model the grasp wrench for one narrow side rectangular object which lies within the grasp radius (20mm$\times$40mm). Here, we consider a total of four forces - two friction cones at the two top corners as shown in Supplementary Figure S\ref{fig:gwh}C and two frictionless point forces at the tip of either end of the grasper. In this case however, there is no external torque acting after the perching maneuver and the ground reaction force helps counter the gravitational forces on the body. With the aforementioned parameters and assuming that all the tip forces at 1 and 2 act towards the center of the object, as shown in the figure, we can compute each wrench force as mentioned in previous section. Specifically, the forces are calculated to be \[f_1 = [0.26 \sqrt{2} ~ 0.26 \sqrt{2}]^T, ~f_2 = [-0.26 \sqrt{2} ~ 0.26 \sqrt{2}]^T,\] \[ f_{31} = f_{41} = [-4.49~ -6.41]^T,~f_{32} = f_{42} = [4.49 -6.41]^T\] The grasper can resist small arbitrary forces in this configuration since the origin lies within the wrench hull. 
	% (shown by the blue dot), $w_{ext} = [0 ~~ -11.12 ~0]^T$, is successfully countered in this case since the grasp wrench hull contains the point $w = [11.12 ~0]^T$ as shown in Supplementary Fig. \ref{fig:gwh}C by the yellow dot. 
	Note that, if the object is larger than the grasp radius, as shown in Supplementary Figure S\ref{fig:gwh}D, the ground reaction force can help stabilize SoBAR's-grasper system after perching, given a flat final orientation. For objects that do not conform to the size but lie within the grasp radius, Supplementary Figure S\ref{fig:gwh}E, the grasper can effectively generate forces to hold on to the perch and resist take-off. This configuration can be modelled as four friction forces, two at the top point of contact and two friction forces at the bottom pints of contact which ensures that the configuration is in force closure. This phenomena is clearly demonstrated in Supplementary Video 5. We successfully employ these insights to demonstrate real-time experiments with SoBAR on objects that ensure successful perches, seen in Sec. 3.3.
	%%%%%%%%%%%%%%%%%%%%%%%%%%%%%%%%%%%%%%%%%%%%%%%%%%%%%%%%%%%%%%%%
	
	\subsection{Perching Drop Tests }
	\label{appendix:perching_drop_tests}
	\subsubsection{Comparison between Rigid, Rigid with Damper and SoBAR Frames}
		To demonstrate the benefit of the soft-body with intrinsic compliance, we have run situational drop experiments in three test scenarios: a) for the rigid frame, b) for rigid frame with damper, c) and for SoBAR at various internal pressures of 70kPa, 138kPa and 208kPa. Each drop test in every scenario was performed three times for an unloaded condition (0g) and a loaded condition (200g) to simulate the weight of onboard electronics and battery. These tests are summarized in the table below and  highlighted in Supplementary Video 6:
	\begin{table}
		\centering
		\caption{Drop Test Results}
		\begin{tabular}{c|c}
			\textbf{Drop Test Scenarios} & \textbf{Success Rate} \\
			\hline
			Rigid Robot Frame (no load) & 3/3 \\
			Rigid Robot Frame (with load) & 0/3 \\
			Rigid Robot Frame  with Damper (no load) & 3/3 \\
			Rigid Robot Frame  with Damper (with load) & 1/3 \\
			SoBAR Frame, All Pressures (no load) & 3/3 \\
			SoBAR Frame (with load) at 70kPa & 3/3 \\
			SoBAR Frame (with load) at 103kPa & 3/3 \\
			SoBAR Frame (with load) at 138kPa & 3/3 
		\end{tabular}
		\label{tab:my_label}
		\vspace{-0.25in}
	\end{table}
	From the results we observe that for the unloaded condition, all variations of the frame perform at a high success rate. This is due to the uniformly distributed mass which gives rise to a non-diagonal stiffness matrix for the robot, diverging from a ideal rigid body's characteristics. However, the second case with the payload attached to the center of gravity of the vehicle, resembles an ideal rigid-body and leads to high rebound velocities with very low impact times making grasping more challenging. 
	With the added off-the shelf foam tape (1/2in depth, V445H Foam Tape) as a damper,
	we noted that the success rate of the loaded rigid robot frame improved. Similarly, the soft robot frame maintains the success rates throughout the variations of its internal pressure and thus stiffness.
	\subsubsection{Performance for Various Objects}
\begin{figure}[b!]
		\vspace{-0.3in}
		\centering
		\includegraphics[trim = 0 55 0 0, clip, width=0.38\textwidth]{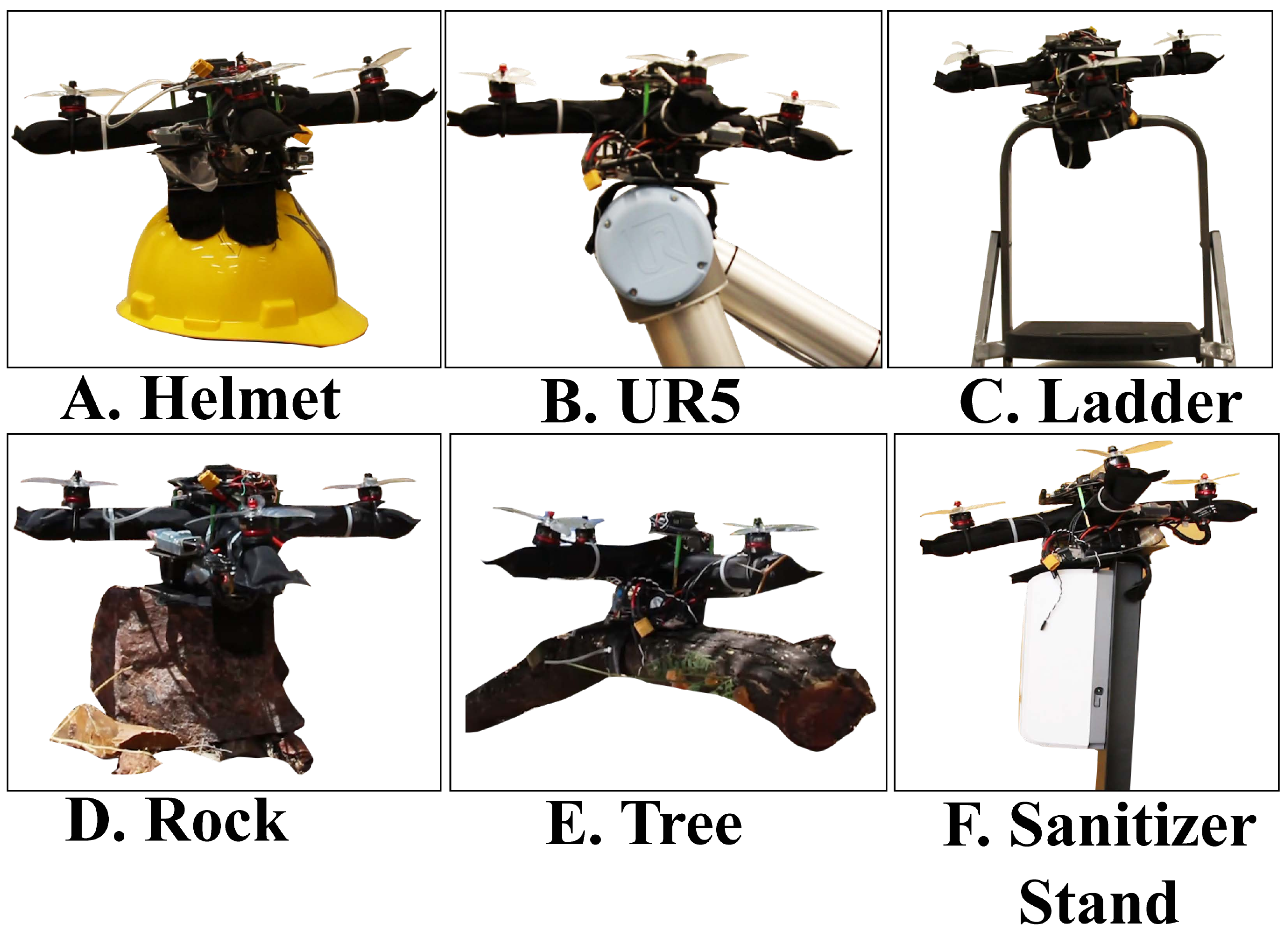}
		% \setlength{\abovecaptionskip}{3pt}
		% \setlenlowcaptionskip}{-10pt}th
		\caption{\textcolor{black}{Perching Drop Test on Various Objects as highlighted in Supplementary Video S6. (A) Helmet (approximate diameter = 22cm) (B) UR5 Robot Arm (approximate diameter = 10cm) (C) Ladder (approximate width = 4cm, height = 2cm) (D) Rock (approximate width = 6.5-8cm, height = 23cm) (E) Sanitizer Stand (approximate width = 13cm, height = 18cm, length = 10cm)}}
		\label{fig:perching_objects}
		% \vspace{-1.5em}
	\end{figure}

	In order to test the system's perching capabilities we selected everyday objects of different sizes and shapes, such as a hard-hat helmet (22cm diameter), edge of a ladder (4cm width, 2cm height), a rock (6.5-8cm width, 23cm height), a tree branch (7.3cm diameter), a joint of a UR-5 robot arm (10cm diameter), and a sanitizer stand (13x18x10cm), highlighting the capabilities of the system to perch onto objects with different shapes and sizes, as seen in Figure S\ref{fig:perching_objects} Supplementary Video 7.
	\subsection{Collision Drop Tests}
	\label{appendix:all_collision_drop_tests}
	In this section, we present the rest of the slow motion screen captures for the collision drop tests that were not shown in Fig. 8. 
	\begin{figure}
		\centering
		\includegraphics[width =0.45
		\textwidth]{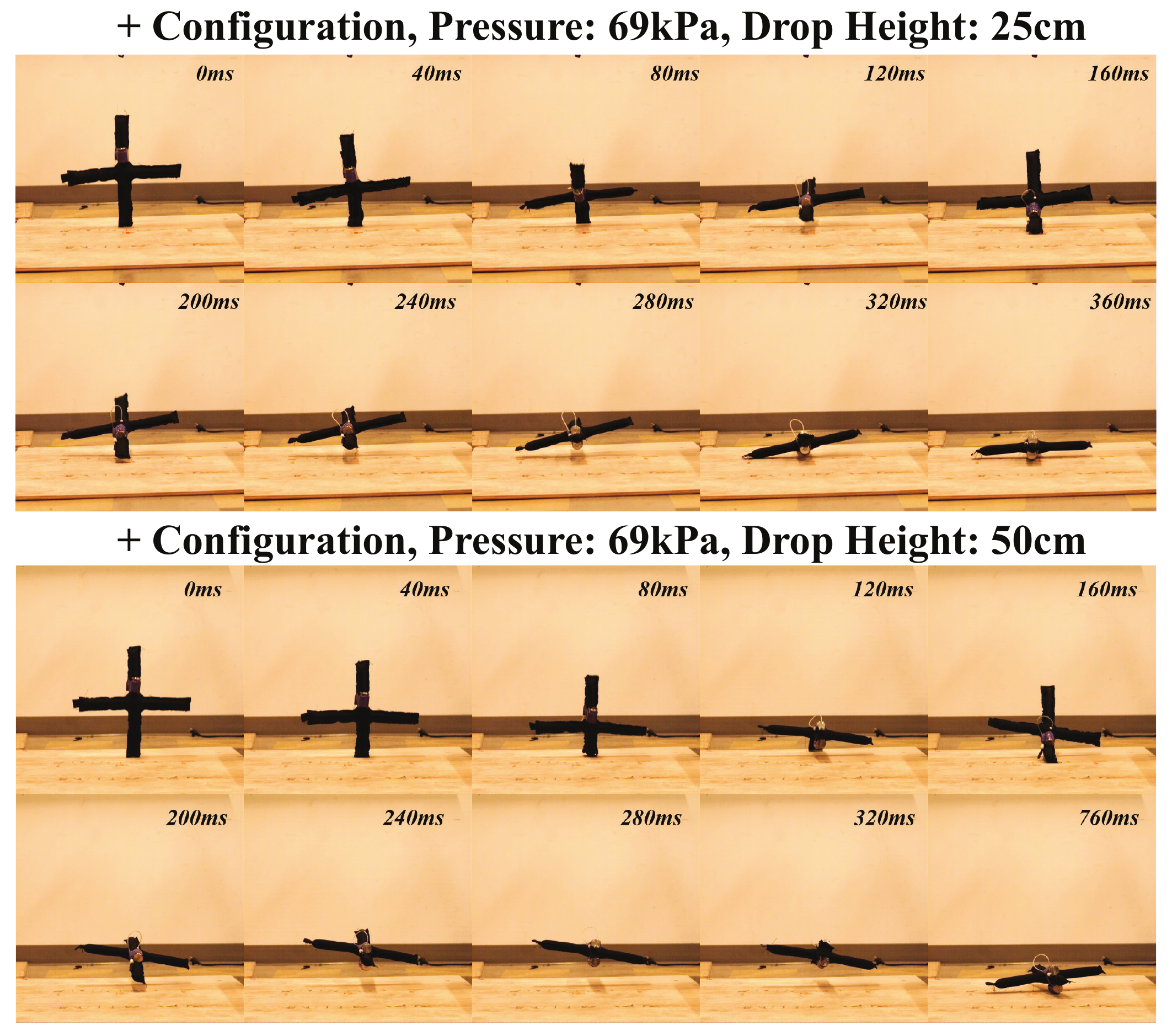}
		\vspace{-0.15in}
		\caption{Collision drop tests for SoBAR's soft-bodied frame in `+' configuration with internal pressures of 69kPa at 25 and 50cm drop heights.}
		% \subfloat[Wrench Hull Obtained]{\includegraphics[width = 0.4\textwidth]{}}
		\label{fig:0_soft_10psi_plus_01}
		\vspace{-0.2in}
	\end{figure}
	\begin{figure}
		\centering
		\includegraphics[width =0.45
		\textwidth]{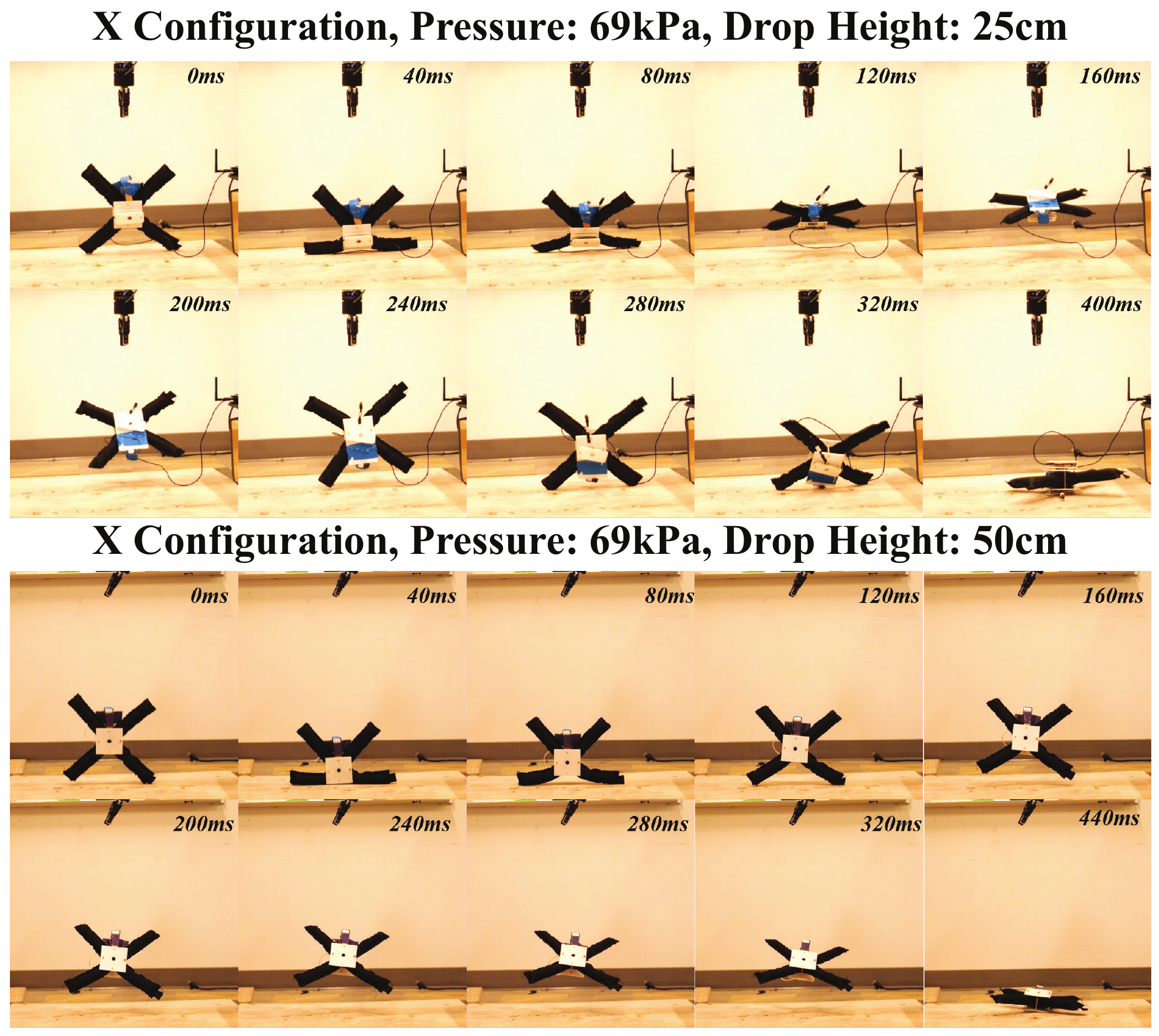}
		\vspace{-0.15in}
		\caption{Collision drop tests for SoBAR's soft-bodied frame in `x' configuration with internal pressures of 69kPa at 25 and 50cm drop heights.}
		% \subfloat[Wrench Hull Obtained]{\includegraphics[width = 0.4\textwidth]{}}
		\label{fig:1_soft_10psi_x_01}
		\vspace{-0.2in}
	\end{figure}
	\begin{figure}[h!]
		\centering
		\includegraphics[width =0.45
		\textwidth]{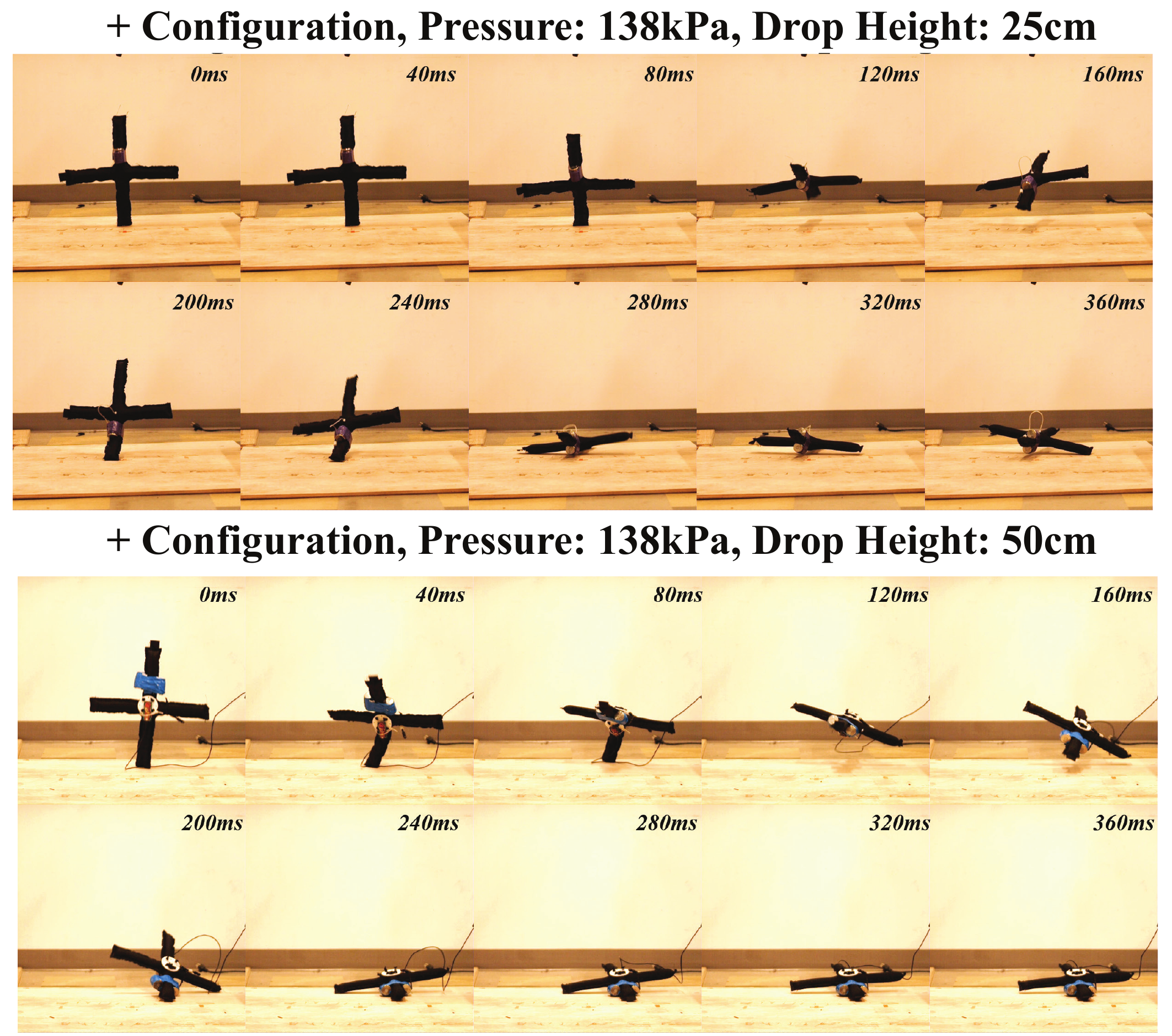}
		\vspace{-0.15in}
		\caption{Collision drop tests for SoBAR's soft-bodied frame in `+' configuration with internal pressures of 138kPa at 25 and 50cm drop heights.}
		% \subfloat[Wrench Hull Obtained]{\includegraphics[width = 0.4\textwidth]{}}
		\label{fig:2_soft_20psi_plus-01}
		\vspace{-0.15in}
	\end{figure}
	\begin{figure}
		\centering
		\includegraphics[width =0.45
		\textwidth]{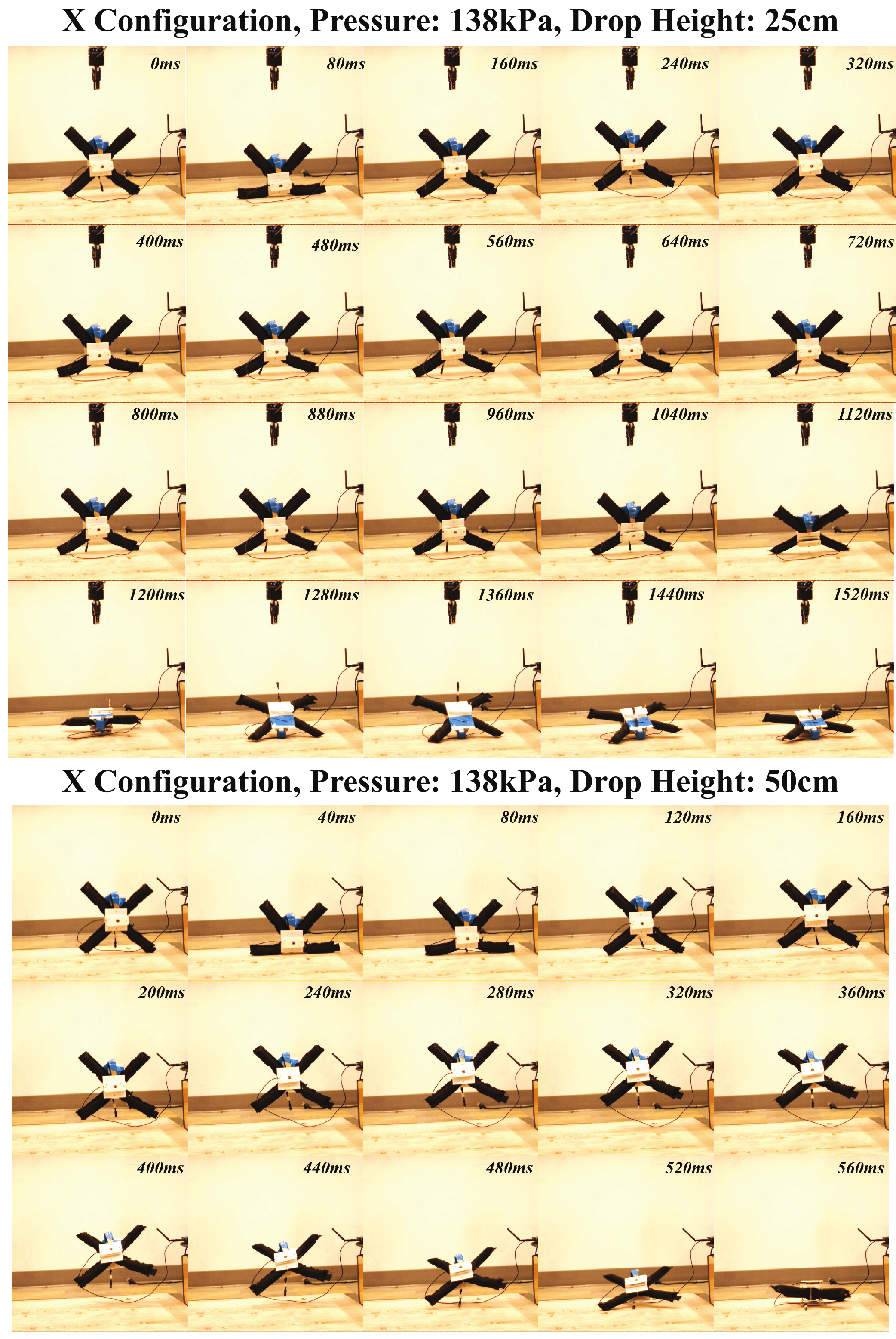}
		\caption{Collision drop tests for SoBAR's soft-bodied frame in `x' configuration with internal pressures of 138kPa at 25 and 50cm drop heights.}
		% \subfloat[Wrench Hull Obtained]{\includegraphics[width = 0.4\textwidth]{}}
		\label{fig:3_soft_20psi_x-01}
		\vspace{-1em}
	\end{figure}
	\begin{figure}
		\centering
		\includegraphics[width =0.45
		\textwidth]{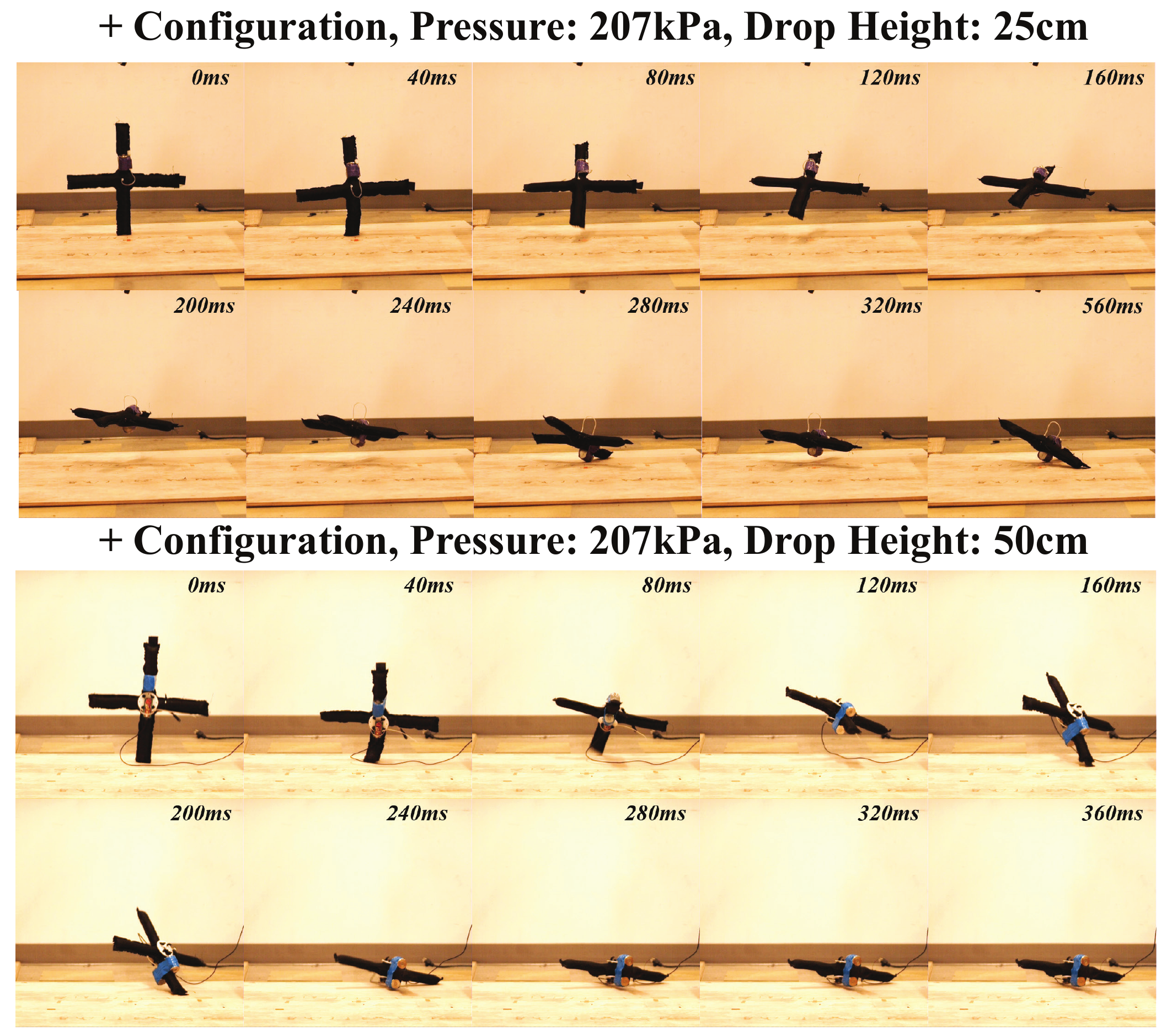}
		\caption{Collision drop tests for SoBAR's soft-bodied frame in `+' configuration with internal pressures of 207kPa at 25 and 50cm drop heights.}
		% \subfloat[Wrench Hull Obtained]{\includegraphics[width = 0.4\textwidth]{}}
		\label{fig:4_soft_30psi_plus-01}
		\vspace{-1em}
	\end{figure}
		The drop tests for the soft-bodied frame at 69kPa are shown in Figs.~S\ref{fig:0_soft_10psi_plus_01} and~S\ref{fig:1_soft_10psi_x_01}. The drop tests for the soft-bodied frame at 138kPa are highlighted in Figs.~S\ref{fig:2_soft_20psi_plus-01} and~S\ref{fig:3_soft_20psi_x-01}. Finally, the drop tests for the soft-bodied frame at 207kPa are presented in Figs.~S\ref{fig:4_soft_30psi_plus-01} and~S\ref{fig:5_soft_30psi_x-01}.
	\textcolor{black}{Please note that the time taken for the drone to drop to the ground varies between each trial, so we chose to show screenshots of 10 different frames. The timing varies slightly because the soft frame might bobble when it hits the ground, so the point it lays completely flat on the ground will vary slightly.}
	\begin{figure}
		\centering
		\includegraphics[width =0.45
		\textwidth]{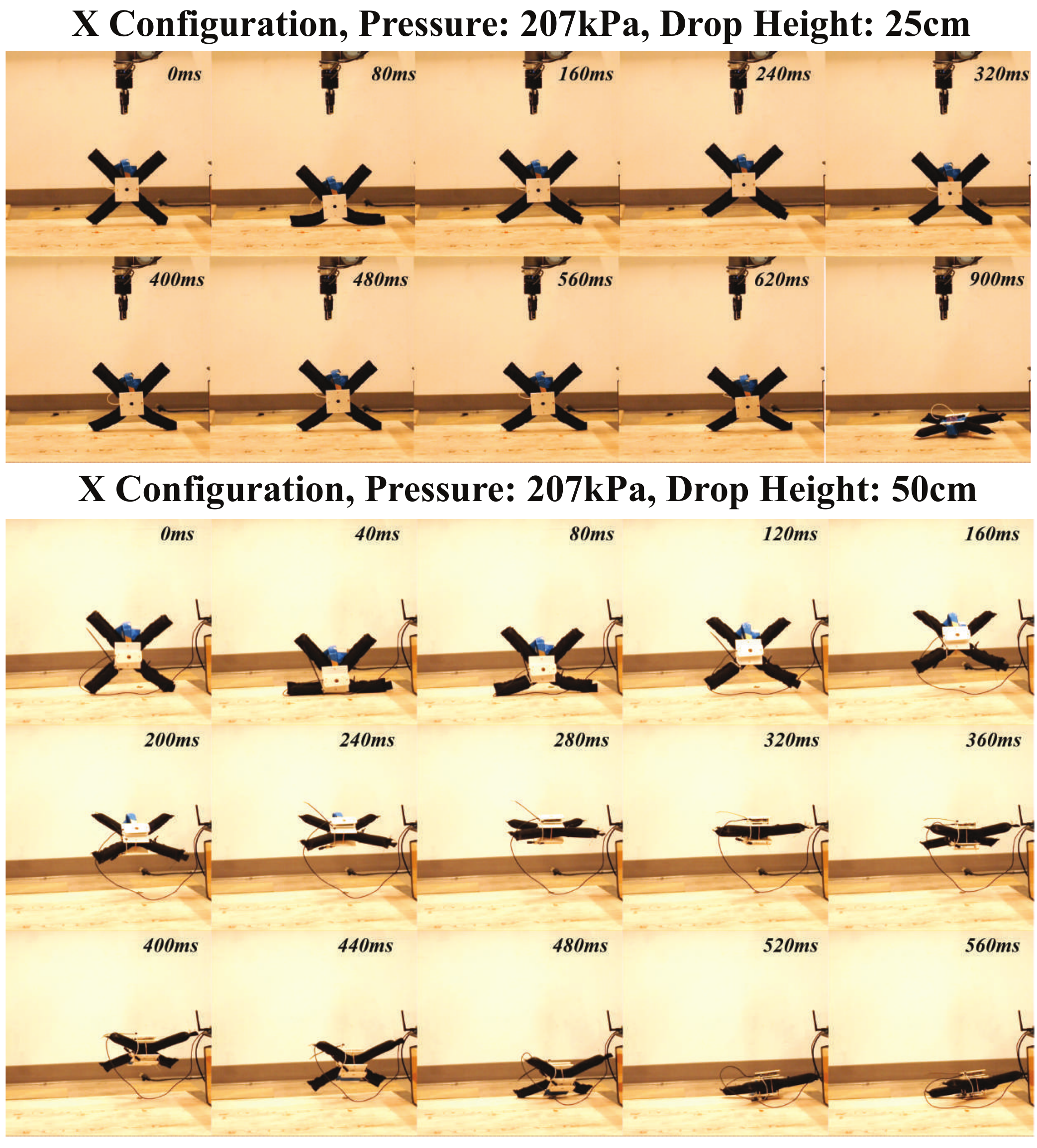}
		\caption{Collision drop tests for SoBAR's soft-bodied frame in `x' configuration with internal pressures of 207kPa at 25 and 50cm drop heights.}
		% \subfloat[Wrench Hull Obtained]{\includegraphics[width = 0.4\textwidth]{}}
		\label{fig:5_soft_30psi_x-01}
		\vspace{-1em}
	\end{figure}
	\begin{table*}[t]
	\centering
	\caption{\textcolor{black}{Comparison of the HFB grasper with other state-of-the-art grippers}}
	\label{tab:HFB}
	\begin{tabular}{p{3cm}p{1cm}p{1cm}p{2cm}p{2cm}p{2cm}p{2cm}p{1cm}}
		\textbf{Gripper Type} & \textbf{Fingers} & \textbf{Weight (Kg)} &  \textbf{Maximum Holding Force (N)} &  \textbf{Power-to-Weight Ratio} & \textbf{Activation} & \textbf{Retraction} & \textbf{Compliant} \\ \hline
		Yale Model \cite{pounds2011grasping} & 4 & 0.49 & 13 & 25.5 & Active & Active & Yes\\ 
		OpenBionics Robotic Gripper \cite{openbionics2020} & 2 & 0.36 & - & - & Active & Active & Yes\\
		Ultrafast Robot Hand \cite{mclaren2019passive} & 3 & 0.5 & 56 & 112 & Passive & Active & Yes\\
		Compliant Bistable Gripper \cite{zhang2019compliant} & 3 & 0.009 & 0.6 & 66.6 & Passive & Active & Yes\\ 
		Passive Claw \cite{stewart2021passive} & 2 & 0.17 & 28 & 164.7 & Passive & Active & Yes \\
		HFB Grasper & 2 & 0.11 & 66 & 600 & Passive & Active & Yes \\ 
		HFB Grasper & 3 & 0.15 & 176 & 1173.3 & Passive & Active & Yes \\ \hline
	\end{tabular}
\vspace{-0.2in}
\end{table*}
	\begin{table}
	\centering
	\caption{\textcolor{black}{Comparison of SoBAR and the HFB grasper with other state-of-the-art perching aerial robots with finger-like graspers. $P_p$ denotes the proportion of the weight of the perching mechanism to the total aerial vehicle weight.~\cite{Meng_2022_perching_review}}}
	\label{tab:HFB2}
	\begin{tabular}{p{2cm}p{1cm}p{1cm}p{1cm}p{1cm}}
		\textbf{Perching Mechanism} & \textbf{$P_p$ $(\%)$} &  \textbf{Stability \cite{Meng_2022_perching_review}} & \textbf{Cylindrical Objects} & \textbf{Planar Objects}  \\ \hline
		Avian-inspired Perching Drone \cite{Doyle2013AvianPercher} & 24 & No & Yes & No  \\
		Bird-inspired Perching Drone \cite{nadan2019bird} & - & - & Yes & No  \\
		Bird-inspired Perching Drone \cite{roderick2017bioinspired} & 33.3 & - & Yes & No  \\
		Passive Perching Fixed-Wing UAV \cite{stewart2021passive} & 5 & Yes & Yes & No \\
		SoBAR with 2 Finger HFB Grasper  & 8.8 & Yes & Yes & Yes  \\\hline
	\end{tabular}
%\vspace{-0.2in}
\end{table}
\begin{table}[t]
	\centering
	\caption{\textcolor{black}{Comparison of SoBAR with other state-of-the-art collision-resilient aerial robots.}}
	\label{tab:collision_comparison}
	\begin{tabular}{p{2cm}p{1cm}p{1cm}p{1cm}p{1cm}}
		\textbf{Vehicle Type} & \textbf{External Protection} & \textbf{Tested impact velocity (m/s) } & \textbf{Maximum impact force (N) } & \textbf{Collision in body-$z$ direction?} \\ \hline
		Quadrotor 'x' & No & 3.13 & 449.46 & No\\
		Quadrotor 'p' & No & 3.13 & 432.89 & No\\
		Euler-spring based \cite{klaptocz2013euler} & Yes & 3.13 & 275 & Yes\\
		Rotorigami \cite{Sareh2018Rotorigami:Rotorcraft} & Yes & 1.2 & 8 & No \\
		Rotorigami \cite{Sareh2018Rotorigami:Rotorcraft} & Yes & 3.13 & - & No\\
		Tensegrity \cite{zha2020collision} & Yes & 6.5 & - & Yes\\
		ARQ \cite{liu2020toward} & Yes & 2.58 & - & No\\
		CRFQ \cite{patnaik2021collision} & Yes & 2.5 & - & No\\
		SoBAR `x' at 207 kPa & No & 3.13 & 52.87 & Yes\\
		SoBAR `p' at 207 kPa & No & 3.13 & 133.31 & Yes\\\hline
	\end{tabular}
\vspace{-0.2in}
\end{table}
	\begin{figure*}[t]
	\centering
	\includegraphics[width = 0.98\textwidth]{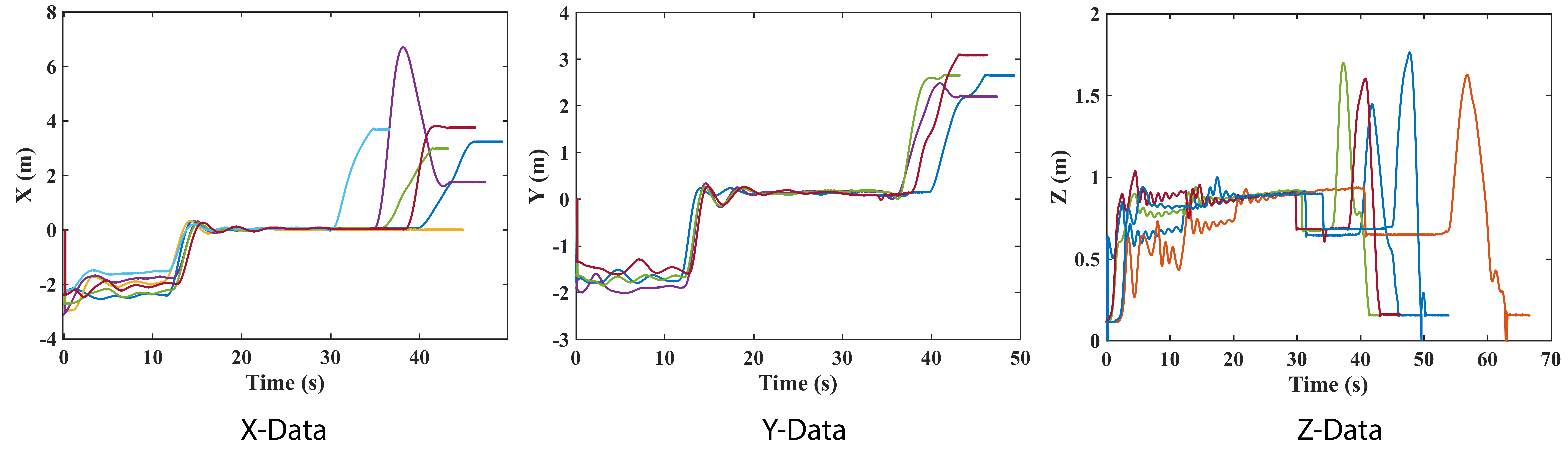}
	\caption{The tracking error for hover during multiple trials of autonomous perching attempts. The tracking error for the hover is around 0.015m in X, 0.031m in Y and 0.051m in Z direction. Moreover, the rise time for X and Y directions is around 2.14s and 1.4s, however the settling time for both is around 8.79s and 12.78s respectively.}
	\label{fig:hover}
\end{figure*}
\begin{figure}
	\centering
	\includegraphics[width = 0.4\textwidth]{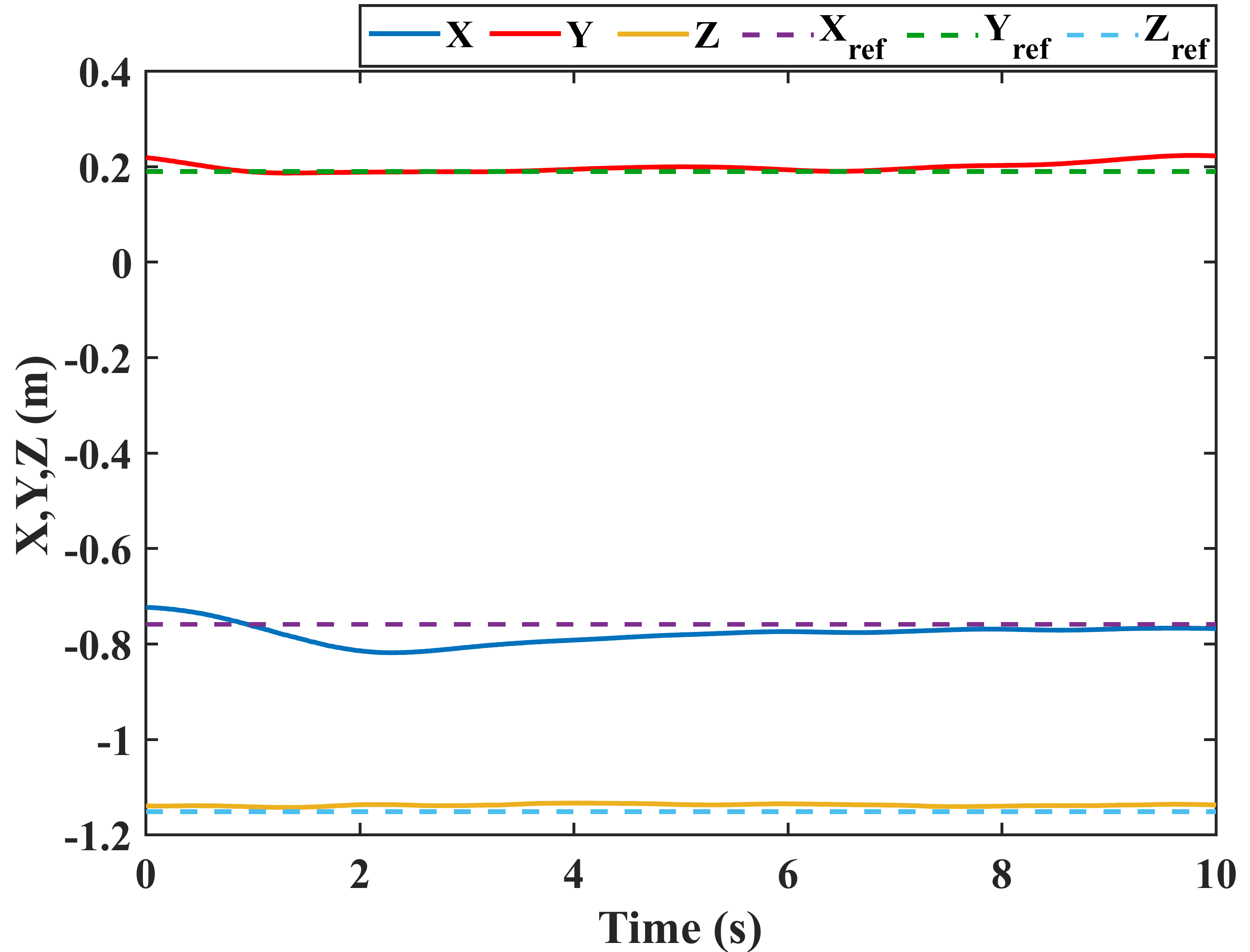}
	\caption{The hover results for a rigid quadrotor (DJIF450) for a 10 sec hover.}
	\label{fig:hover_rigid}
	\vspace{-0.3in}
\end{figure}
	\subsection{Comparison with Other Perching Aerial Robots}
		In this section we compare the HFB Grasper with the state-of-art perching methods available \cite{Meng_2022_perching_review}. In Table \ref{tab:HFB}, we present the comparison between different grippers available and their holding force, against the holding force capability of the presented HFB grasper. From Table \ref{tab:HFB}, we notice that the HFB Grasper has a power-to-weight ratio of 600 and 1173, for the two-fingered and three-fingered version, respectively. This is significantly higher than the comparable graspers used for perching, as discussed in Meng et. al~\cite{Meng_2022_perching_review}.\\
		In Table \ref{tab:HFB2}, we present the stability of the post perching state of various aerial robots capable of perching with finger-like graspers. The variable $P_p$~\cite{Meng_2022_perching_review}, highlights the proportion of the grasper weight over the weight of the quadrotor. From the table, we notice that in comparison to similar perching aerial robots, SoBAR's lightweight grasper leads to the lowest $P_p$ comparatively. In addition, SoBAR is capable of perching on irregular objects, cylindrical objects and also on planar surfaces. Furthermore, the perching is stable in various orientations.	
	\subsection{Comparison with Other Collision Resilient Aerial Robots}
		In this section we provide a comparison between some existing collision resilient quadrotors in terms of impact velocity and maximum impact force, highlighted in Table \ref{tab:collision_comparison}. It is noted that only a few collision resilient quadrotors in literature measure the impact force endured by the center of the body of the system. From Table~\ref{tab:collision_comparison} we note that a standardized quadrotor with just propeller guards, tested at 3.13 m/s experiences an impact force of 450N. When this is compared to SoBAR, with a soft-body, mitigates 11.2x the impact force experienced by the flight controller. Furthermore, SoBAR is capable of absorbing impact in the body-z direction as well, in order to assist with perching, without needing further mechanical additions.
	\subsection{Low-level performance of SoBAR}
		In this section, we present the results for tracking performance of SoBAR during a hover condition. As described in Sec 2.4, due to beam deflection, the thrust at any given instant generated by each of the four motor-propeller pairs will have net residuals affecting horizontal drift. In free flight and near-hover condition (when there is no external physical force acting on the drone), the motors are friction fit on the arms such that there is no rotation about the $x_{arm}$ axis, therefore just leaving the deflection with respect to the $y_{arm}$ axis. The effective drift force in the horizontal plane can be calculated by:
		\begin{equation}
			\Delta_f = [F_1 \sin \theta_1 - F_{3} \sin \theta_{3}~~~ F_2 \sin \theta_2 - F_{4} \sin \theta_{4}~~~0]^T
		\end{equation}
		Considering the flights were conducted for the soft body greater than a pressure of 137.89 kPa, $\theta$ roughly corresponds to $9.92\pm1.28^o$. With a maximum thrust of $10\pm0.5$ N for each arm, 
		$0\text{N} < \|\Delta_f{|_{137.89kPa}}\| < 0.62$N.
		At 207kPa with deflection angle around $5.80\pm1.19^o$, 
		$0\text{N} < \|\Delta_f{|_{207kPa}}\| < 0.51$N.
		% With a mass of 1.4kg, this corresponds to a maximum drifting acceleration of $\pm$0.44m/s$^2$ and $\pm$0.36m/$s^2$ respectively for 137.89kPa and 207kPa. 
		During flights, the pitch, roll gains are high enough to overcome the horizontal drift as is shown in our perching experiments, attached as Supplementary Video 5 and in the Fig. \ref{fig:hover}. As also seen in Fig. S2,  the tracking error for the hover is around 0.015m in X, 0.031m in Y and and 0.051m in Z direction with the setpoint as $x_h = [0.04~0.17~0.85]^T$ which represents a setpoint directly above the target perching location $x_{obj} = [0.04~0.17~0.65]^T$, which corresponds to a location on the perching object. The rise time for X and Y directions is around 2.14s and 1.4s, how ever the settling time for both is around 8.79s and 12.78s respectively due to the slightly varying beam stiffness for individual arms (effect of fabrication error). The tracking error for a rigid conventional DJIF450 quadrotor was observed to be around 0.02m, 0.0084m and 0.013m respectively for the X,Y and Z directions, as shown in Fig. \ref{fig:hover_rigid}. Hence the tracking performance of SoBAR is comparable to that of the rigid designs with similar dimensions. The flight time and the tracking errors for SoBAR with its chosen electronic components are summarized in the Table \ref{tab:flightchar}.
	\begin{table}
		\centering
		\caption{\textcolor{black}{Flight Specifications for SoBAR}}
		\label{tab:flightchar}
		\begin{tabular}{p{4cm}p{2cm}}
			\textbf{Parameter} & \textbf{Value} \\ \hline
			Battery & 4S 2200mAh \\
			Motors & 2300kV \\
			Propellers & 5in \\
			Total weight for hover test & 864g \\
			Flight time & 1min30s\\
			Tracking error & [0.015m~ 0.031m~ 0.051m]$^T$ \\
			\hline
		\end{tabular}
	\end{table}
	
	\subsection{Slip Detection}
\begin{figure}
		\centering
		\includegraphics[width = 0.48\textwidth]{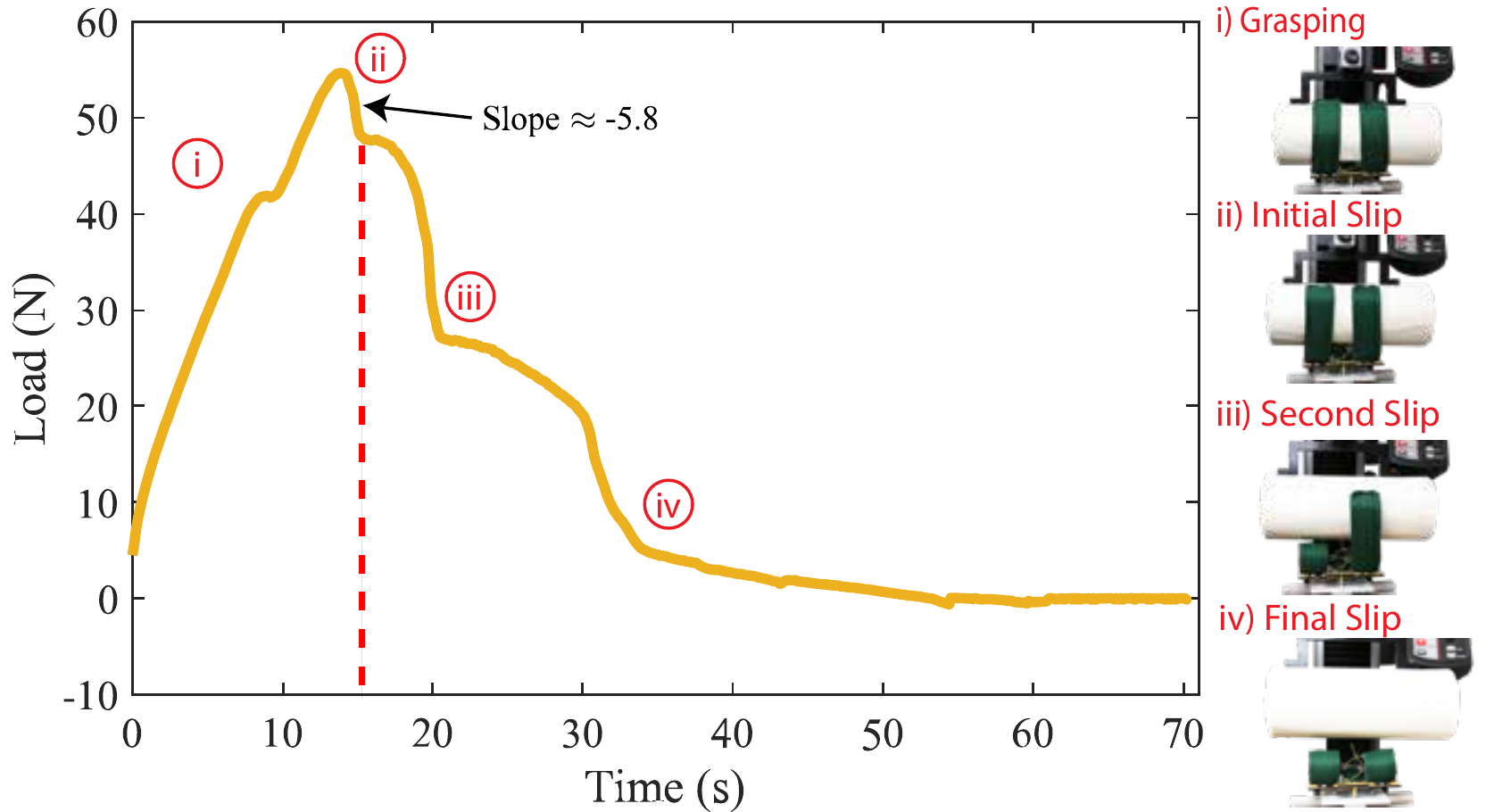}
		\caption{Slip Detection Methodology Demonstration  on a 55mm Diameter Object with a 2 Finger Grasper}
		\label{fig:slip}
	\end{figure}

		This section describes how slip was detected for characterizing the HFB grasper. Initially when the UTM pulls on the object, the grasper exerts forces to retain the grasp and we see the load increase gradually. However, as the UTM continuously increases the pulling force, at one point (indicated by the arrow in Fig. \ref{fig:slip}) the grasping force decreases drastically indicating a slip condition for the right half of one finger. The next steep negative slope shows the release by the other half of the finger and finally the entire 2 fingers are released. To characterize the maximum grasping force of the grasper on that particular object, we note the maximum value observed on the UTM. Therefore, slip was detected when the UTM data showed a negative of slope angle greater than $80^o$ or -5.8.

		We present an example of the grasping force test with the universal tensile testing machine (UTM) as a graph in Fig. \ref{fig:slip}. Further, the slip detection video at \url{https://youtu.be/JgrYh7fI_HM}), shows the various stages from grasping to multiple slip locations. 
	
	\subsection{Perching Demonstration Alongside Wall Collision}
	In this section, we show the capability of SoBAR to successfully perch while at the same time colliding with a wall. Keeping safety requirements in view and to avoid crashes, we have designed another situational experiment between the rigid robot frame, rigid robot frame with damper, and SoBAR frame at various pressures. The frames were hand-launched towards the perch, placed next to the wall at a distance of 0.2m as shown in Supplementary Video 8. It is to be noted that the manual launching velocity and angle was aimed to be similar for all trials.
	
	In the Table \ref{tab:wallcol} below, the various success rates for the all frames can be seen. The rigid frame with and without the damper failed to perch in all trials. This is attributed to two main reasons. First, the rebound velocity from the collision with the wall is high enough causing the vehicle to slip sideways before the perch is completed with the engaged gripper. In the second case, the approach angle is such that the gripper is not aligned with the target due to the non-conforming nature of the rigid arm. The SoBAR frame, at higher stiffness (at internal pressures of 138kPa and 103kPa), also failed on several of the trials. What was interesting is the deformability of the SoBAR frame at 70kPa which allowed reconfiguration of its arm upon impact and successfully perch. However, the recovery seen in the first trial is relatively easier than the second where the arm is squished between the perching target and the wall, making it difficult to recover.
	\begin{table}
		\centering
		\caption{Wall Collision Results}
		\begin{tabular}{c c}
			\textbf{Wall Collision and Perching Test Scenarios} & \textbf{Success Rate} \\
			\hline 
			Rigid Robot Frame & 0/3 \\
			Rigid Robot Frame with Damper & 0/3 \\
			SoBAR Frame at 70kPa & 3/3 \\
			SoBAR Frame at 103kPa & 1/3 \\
			SoBAR Frame at 138kPa & 0/3 \\
			\hline
		\end{tabular}
		\label{tab:wallcol}
	\end{table}
	
%TC:endignore

%%%%%%%%%%%%%%%%%%%%%%%%%%%%%%%%%%%%%%%%%%%%%%%%%%%%%%%%%%%%%%%%%%%%%%%%%%%%%%%%%%%%%%%%%%%%%%%%%%%%%%%%%%%%%%%%%%%%%%%%%%%%%%%%%%%%%%%%%%%%%%%%%%%%%%%%%%%%%%%%%%%%%%%%%%%%%%%%%%%%%%%%%%%%%%%%%%%%%%%%%%%%%%%%%%%%%%%%%%%%%%%%%%%%%%%%%%%%%%%%

\end{document}